\documentclass{article}

    \usepackage[final, nonatbib]{neurips_2021}

\usepackage[utf8]{inputenc} 
\usepackage[T1]{fontenc}    
\usepackage[hidelinks]{hyperref}       
\usepackage{url}            
\usepackage{booktabs}       
\usepackage{amsfonts}       
\usepackage{nicefrac}       
\usepackage{microtype}      
\usepackage{xcolor}         
\usepackage{graphicx}
\usepackage{flafter}
\usepackage{subcaption}
\usepackage{amsmath}
\usepackage{array}
\usepackage{float}
\usepackage{soul}
\usepackage{rotating}
\usepackage{adjustbox}
\usepackage{tikz}
\usetikzlibrary{shapes.geometric, arrows.meta, positioning}
\graphicspath{{images/}}

\title{
ROS-LLM: A ROS framework for embodied AI with task feedback and structured reasoning
}

\author{
Christopher E. Mower\thanks{Huawei Noah's Ark Lab, London, UK},~ 
Yuhui Wan\footnotemark[1]~ \thanks{University of Leeds, Leeds, UK},~ 
Hongzhan Yu\footnotemark[1],~
Antoine Grosnit\footnotemark[1]~ \thanks{Technical University of Darmstadt, Darmstadt, Germany},
\\ 
{\bfseries
Jonas Gonzalez-Billandon\footnotemark[1],~
Matthieu Zimmer\footnotemark[1],~
Jinlong Wang\thanks{East China Normal University, Shanghai, China},~ 
Xinyu Zhang\footnotemark[4],~
Yao Zhao\thanks{Huawei Technologies, Hangzhou, China}, 
}\\
{\bfseries
Anbang Zhai\footnotemark[5],~
Puze Liu\footnotemark[3],~
Daniel Palenicek\footnotemark[3],~
Davide Tateo\footnotemark[3],~
Cesar Cadena\thanks{ETH Zurich, Zurich, Switzerland},~ 
Marco Hutter\footnotemark[6],~ 
}\\
{\bfseries
Jan Peters\footnotemark[3],~ 
Guangjian Tian\thanks{Huawei Noah's Ark Lab, Hong Kong, China},~ 
Yuzheng Zhuang\thanks{Huawei Noah's Ark Lab, Shenzhen, China}, 
Kun Shao\footnotemark[1],~
Xingyue Quan\footnotemark[8],~
Jianye Hao\footnotemark[8],~
}\\
{\bfseries
Jun Wang\thanks{University College London, London, UK},~ 
Haitham Bou-Ammar\footnotemark[1]~ \footnotemark[9]
}
}

\begin{document}

\maketitle

\setcounter{footnote}{0} 

\begin{abstract}
We present a framework for intuitive robot programming by non-experts, leveraging natural language prompts and contextual information from the Robot Operating System (ROS). 
Our system integrates large language models (LLMs), 
enabling non-experts to articulate task requirements to the system through a chat interface. 
Key features of the framework include:
integration of ROS with an AI agent connected to a plethora of open-source and commercial LLMs,
automatic extraction of a behavior from the LLM output and execution of ROS actions/services,
support for three behavior modes (sequence, behavior tree, state machine),
imitation learning for adding new robot actions to the library of possible actions, 
and
LLM reflection via human and environment feedback. 
Extensive experiments validate the framework, showcasing 
robustness, scalability, and versatility in diverse scenarios, including long-horizon tasks, tabletop rearrangements, and remote supervisory control. To facilitate the adoption of our framework and support the reproduction of our results, we have made our code open-source. You can access it at: \href{https://github.com/huawei-noah/HEBO/tree/master/ROSLLM}{\textcolor{blue}{\textbf{ROS-LLM-Code}}}.
\end{abstract}

\begin{figure}[h]
    \centering
    \includegraphics[width=\textwidth]{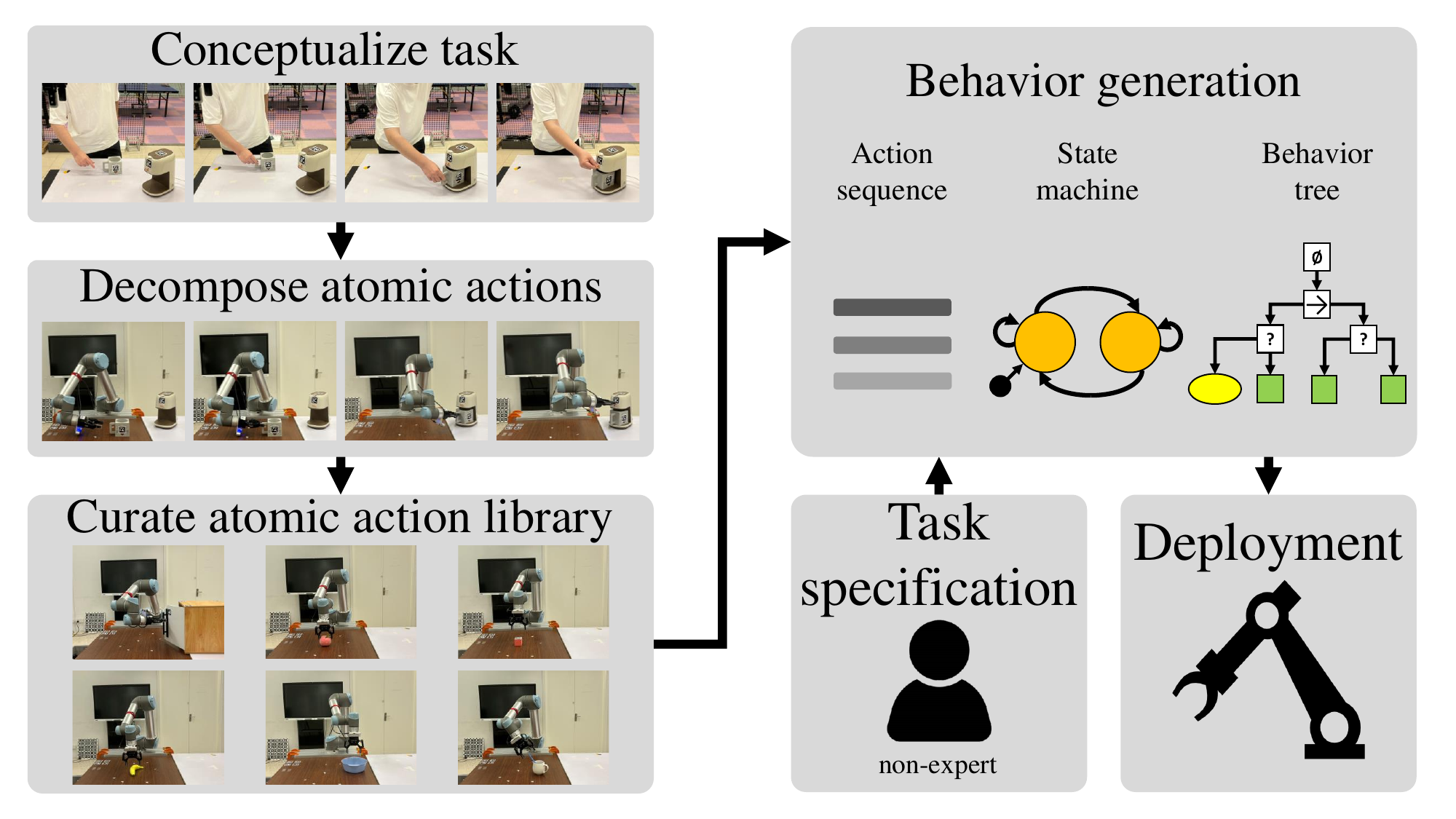}
    \caption{
    Overview of a typical robotics development workflow.
    }
    \label{fig:robotics-dev-overview}
\end{figure}

\section{Introduction}

When developing a robotic system, whether it be in industry or research, typically roboticists follow a similar development cycle, that we summarize in Figure~\ref{fig:robotics-dev-overview}.
Initially, a task is \textit{conceptualized}, for example ``make me a coffee''.
Formally, we will refer to such a task for the robot as an \textit{action}.
In our case, we assume actions from domestic scenarios, 
however, the action could be inspired by any other application 
(e.g. healthcare, construction, space). 

The conceptualized task, or action, is broken down into several sub-tasks, e.g. ``reach'', ``pick'', ``switch'', etc. 
Throughout this paper, we refer to these  sub-tasks as \textit{atomic actions} since a complete action, such as ``make me coffee'',
comprises several components within a larger sequence or combination of sub-tasks.
For each atomic action, a low-level policy representation can be obtained using common approaches such as:
reinforcement learning (RL), 
imitation learning, or
optimal control  (e.g. model-predictive control).
Note that, in our case, atomic actions can also include perception sequences (e.g. ``locate mug''). 

As we conceptualize more tasks, we subsequently develop more and more atomic actions that we can curate in a library.
Each atomic action is ultimately a function: an input variable and parameters that are mapped to a return value.
The representation of an atomic action could be simple, for instance, ``open gripper'' that sends an ``open'' signal to a parallel gripper attached to a robot arm which then returns the final width from the gripper.
The representation could take other forms, for example 
a planner and feedback controller, 
a neural network policy trained using RL or imitation learning for the end-effector that feeds into an inverse-kinematic controller to compute target joint states.
The planner/controller parameters for a given formulation or the neural network weights can be stored in memory and considered as the atomic action representation in the atomic action library.
This library of atomic actions can be in the form of a code API or, if the Robot Operating System (ROS) is being used, the library can be a list of ROS actions and services. 
Textual descriptions for each atomic action are always assumed to be provided, i.e. documentation.

A library of atomic actions is collected and maintained by experts:
such as the robot manufacturer 
(e.g. KUKA, ABB, Clearpath) 
or 
companies providing integration services
(e.g. Covariant, Berkshire Grey, PickNik Robotics).
Interestingly, the robot system can perform unseen tasks by combining some atomic actions. Those novel tasks would typically be specified to the manufacturer/integrator by non-expert customers.
Common behavior representations are sequences of atomic actions, state machines, or behavior trees. 
Given a combination of atomic actions in one of these form, the robot system can then be deployed.
We note that, despite this library of atomic actions,
teams of expert robot engineers often need to manually develop and deploy the proper combination of atomic actions.

The approach shown in Figure \ref{fig:robotics-dev-overview} and described above has proven effective in many applications such as manufacturing, logistics, and inspection, e.g.~\cite{IOVINO2022104096, Everett94}.
These applications where robotics have been successfully applied generally involve repetitive situations.
Furthermore, these scenarios typically involve task specifications that are unlikely to change frequently. 
For instance, a vehicle manufacturer shall probably not alter their process once the production line is set up and operational.
However, if a new behavior or action is needed for the system, revising or rewriting the system workflow will require the work a team of expert robotics engineers.

The development and adoption of robotics in applications such as domestic use, healthcare, and construction has been slower than in the applications mentioned above.
This can be explained by the challenges needed to address such as  frequently changing and time-dependent task specifications,
interaction and collaboration with non-experts, and 
the need to extend the range of capabilities of the system via input from non-experts.
For these reasons, it is unreasonable to assume teams of experts (expensive in time and money) will be on-hand to update the capabilities of the system at will and 
within reasonable time limits.

In general, robot systems are designed in a modular way to allow users to build their own frameworks by easily integrating and modifying existing processes. 
The most well-known framework used in research and industry is the 
Robot Operating System (ROS)~\cite{quigley2009ros, ros2}. 
Other examples include the Lightweight Communications and Marshaling (LCM) project~\cite{Huang10} and the Open Robot Control Software (Orocos) project~\cite{Bruyninckx2001}.
The ROS framework provides a well-established ecosystem of packages and libraries that are ready to use and integrated with many common robot systems (e.g. Universal Robots, Robotiq, Clearpath Robotics).
Some of most widely used ROS libraries and packages include the TF library \cite{Foote13}, MoveIt~\cite{coleman2014reducing}, and the Navigation-stack.
The ROS ecosystem of contributors additionally contains many packages for many important requirements such as simulation~\cite{Koenig-2004-394, mower2023ros}, kinematic modeling~\cite{kdl-url, Beeson15}, and planning and control~\cite{huber2024lbrstackros2python, Mower23, Chitta2017}.
Thus, ROS offers many packages providing useful functionalities for both research and commercial applications. 
These packages include valuable data structures, control interfaces, inverse kinematics (IK) and motion planning tools, perception utilities, and various visualizers. 
Additionally, with new tools such as the \texttt{BehaviorTree.ROS} library, ROS actions and services enable the generalization of a wide variety of capabilities required by robot systems into a unified execution framework.

In our work, we aim to provide a framework that addresses the needs and limitations described above.
We argue that a key bottleneck in the current robot development workflow is the reliance on teams of robot experts.
Therefore, our goal is to develop a system that enables 
\textit{non-experts to program robots}, i.e. provide the robot with new action compositions and extend the capabilities of the system through demonstration.
Given its popularity and the general need for modularity, we integrate our framework with ROS.

Many advancements have been made in robotics and machine learning. 
Notably, the field of AI has made significant advances recently in natural language processing, due primarily to the development of transformer models~\cite{vaswani2017attention} and subsequently large language models~\cite{openai2024gpt4, guo2024deepseekcoder, touvron2023llama}.
These language models have been shown to exhibit remarkable performance across a broad spectrum of tasks
such as
generating code~\cite{chen2021evaluating},
solving advanced mathematics problems~\cite{frieder2023mathematical, Trinh2024},
generating valid chain-of-thoughts~\cite{Wei2022,tutunov2023large}, providing
search results~\cite{hagrid}, 
assisting in medical diagnosis~\cite{Singhal2023}, and others.

Language is a natural tool humans use daily to communicate their goals to one another.
With recent advancements in language models, there  is increasing interest in using these models to map natural language to a sequence of robot actions.
By expressing  task requirements in natural language and providing contextual information retrieved from the robot's environment, non-experts can effectively convey their intentions to robotic systems. 

Leveraging natural language, our method generates behavior representations tailored to the task specified by a non-expert user, thereby alleviating the burden on expert engineers and enabling rapid prototyping and deployment of robotic systems.

Moreover, since a language model will likely not be capable enough to replace a team of experts, 
our framework enables continuous learning and adaptation, 
facilitated by iterative feedback loops.
Also, in-order to enable non-experts to update the library of atomic actions via imitation learning, we develop an interface for them to provide several demonstrations.
Through teleoperation or kinesthetic teaching, users can augment the system's atomic action library with skills, enriching its repertoire of behaviors. 
The human feedback provides the system with the ability to correct its past mistakes and to modify task objectives.

In this paper, we make the following contributions.

\begin{itemize}

\item Introduction of a framework that enables non-experts to intuitively program robots using natural language prompts and contextual information retrieved from a ROS environment (i.e. mapping sensor readings to text).
\item Integration of an open-source language model with a comprehensive atomic action library, facilitating the composition of action sequences for complex long-horizon tasks that are adaptive to dynamic environments and human feedback.
\item Development of a real-world robot setup within a kitchen-like environment and performing various experiments that provide important lessons on prompting strategies to improve the performance of open-source language models.
\item Demonstration of the system in handling diverse scenarios, showcasing its robustness and versatility in human-robot collaboration, along with the provision of key insights for using open-source LLMs in robotics applications. Furthermore, we perform a long-distance supervisory control user-study where an operator in Europe controls a robot in Asia to perform a tabletop rearrangement task. Additionally, we describe two use-cases of our system being developed/deployed in systems in the wild.
\end{itemize}

\section{Problem formulation}

As described in the introduction, we can think of an atomic action as a single task that the system can perform. 
Formally, we frame a single task as a Markov Decision Process (MDP) characterized by
the tuple $\langle \mathcal{S}, \mathcal{A}, r, \mathcal{P}, \gamma \rangle$
where 
$\mathcal{S}$ is the state space,
$\mathcal{A}$ is the action space,
$r: \mathcal{S} \times \mathcal{A} \to \mathbb{R}$ is a reward function defined for any  state $s\in\mathcal{S}$ and any action $a\in\mathcal{A}$, 
$\mathcal{P}(s_{t+1} | s_t, a_t)$ is a transition probability distribution, 
$t$ is a discrete time step, and
the scalar $0 < \gamma \leq 1$ is a discount factor. 

In contrast, to the standard MDP formulation, we also assume access to a failure flag $f$ that is returned on termination of the MDP, i.e. task completion.
The failure flag indicates whether the desired task was completed successfully or not, i.e. $f=0$ indicates success, $f=1$ otherwise.
For example, if the task is for a robot arm to reach to a target, then at termination $f=0$ means the target was acquired, and $f=1$ implies that the robot finished in a configuration far from the target.
Thus, our modified MDP is denoted by $\langle\mathcal{S}, \mathcal{A}, r, f, \mathcal{P}, \gamma \rangle$.

In single-task RL, an AI agent generally learns a policy $\mu(a | o)$  choosing an action $a$ given an observation $o$ (e.g. from an image) of the state $s$.
The agent's objective is to determine the sequence of actions that maximizes the expected return $\mathbb{E}\left[\sum_t \gamma^t r(s_t, a_t)\right]$.
The specific task and its associated rewards are determined by the reward function $r$.
In our case, we assume access to an atomic action library that corresponds to having a set of $N$ pre-trained or pre-defined policies $\{\widehat{\mu}_{i}\}_{i = 1}^N$ ready to use, each based on an underlying modified MDP.
Note, we use hat $\widehat{\mu}$ to denote that the policy is trained/defined.

Humans exhibit remarkable proficiency in synthesizing complex behaviors by composing various known skills. 
With regard to a robotic system, this process involves the composition of distinct policies 
that are executed following a certain structure, such as some appropriate sequential order, or corresponding to a data structure like a behavior tree.
In situations when there is access to a team of experts,
it is reasonable to assume that they are capable to define some reward function that measures appropriate compositions of atomic actions, i.e. behaviors. 
With access to experts, we can reasonably assume some reward or fitness function that specifies an appropriate composition of these policies, 
in which case we could explore methods based on hierarchical reinforcement learning~\cite{Barto2003}.
However, in our case, absent expert guidance, the robotic system must rely on environmental observations $o$ and non-expert human input $h$ to guide policy selection.
Also, we assume the input from the human is given by text.

Assuming that an appropriate data representation (e.g. a sequence or a tree) has been selected, 
it denotes a specific configuration of atomic action composition by $b$ and terms this as a \textit{behavior}.
For example, in the case of a sequence, given an atomic action library with $N=3$ policies, we could choose $b = \{3, 1, 2, 1\}$ meaning we should start executing policy $\widehat{\mu}_3$ first, then $\widehat{\mu}_1$, $\widehat{\mu}_2$, and finally $\widehat{\mu}_1$.
As mentioned above, this choice of composition $b$ is based on the observation $o$ and human input $h$.
Thus, we define the mapping $\pi(b | o, h)$ as a \textit{behavior policy} - this is our AI agent.

When a behavior $b$ is chosen by the behavior policy $\pi$, 
it may be the case that one of these policies fails or all may succeed. 
In the case of failure, it is typically not worthwhile to proceed.
For example, if a robot reaches to grab a mug but knocks it over when attempting to grasp due to misalignment, 
then it would not make sense to continue to try to place the mug.
Thus, in this case, we consider that the result of the behavior $b$ is given by $\bar{f} = 1$. 
When all the atomic actions complete successfully, then the result of the behavior $b$ is $\bar{f} = 0$.
Note, the bar $\bar{f}$ indicates the failure result of the behavior time step $\tau$.
Another way to conceptualize this idea is that the final atomic action failure flag is given as the failure flag of the behavior $b$.


To enhance the selection of optimal behaviors $b$ and improve the behavior policy 
$\pi$, 
the agent requires access to a measure of goodness. 
As previously noted, we refrain from relying on domain experts to formulate new reward functions tailored to specific tasks. 
Under our premise, where access to a failure indicator $f$ is assumed, a clear definition is needed. 
In our current work, we adopt the simplifying assumption that the failure indicator can be hard-coded. 
For instance, during the execution of the ``grasp object'' atomic action, 
the final gripper width serves as an indicator: 
zero width implies failure to grasp the object, while a width greater than zero indicates success. 
This approach is constrained by the reliance on straightforward rules or expert input to establish the failure indicator. 
Recent work, however, suggest promise in training neural networks to classify multimodal failures in manipulation tasks~\cite{Inceoglu24}.
Of interest in our future work is to explore the use of multi-modal transformer networks (e.g. vision-language models) to model the failure indicator $f$.
Ideally, therefore, the agent should choose $b$ to maximize success and also minimize time. 
Since, by assumption, we have access to the failure flags $\widehat{f}$ we define for the behavior policy $\pi$ a return function
\begin{equation}
    \label{eq:rosllm-return}
    R_\pi := \mathbb{E}\bigg[ \sum_\tau -\beta^\tau \Big(1 + \widehat{f}_\tau\Big) \bigg]
\end{equation}
where 
$\tau$ is a time step for the behavior policy $\pi$, and
$0 < \beta \leq 1$ is a scalar discount factor.
Notice that we add 1 to the the failure flag $\widehat{f}_\tau$.
This is required to indicate to the agent that as few atomic actions should be used to construct a behavior as possible. 
Our goal in this work, is to develop a framework that is capable of integrating with real systems that addresses the above novel formulation.

\section{The ROS-LLM Framework}

\begin{figure}[t]
    \centering
    \includegraphics[width=\textwidth]{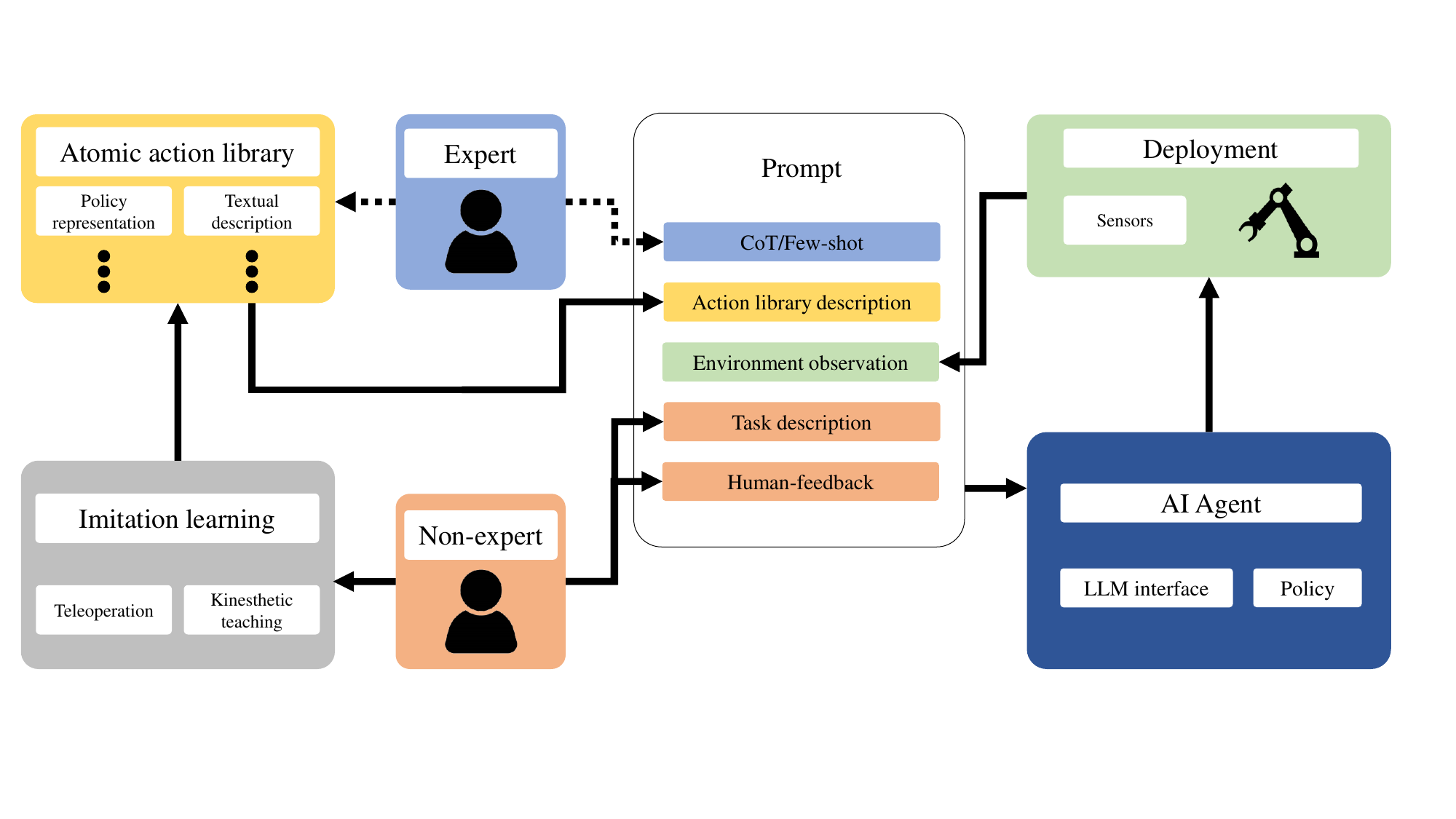}
    \caption{Our proposed ROS-LLM framework overview illustrates the integration of several components. In this figure, the dashed lines denote elements that are only introduced once, such as the initial version of the atomic action library and CoT/Few-shot prompts by an expert.}
    \label{fig:rosllm-framework}
\end{figure}

In this section, we provide an overview of our framework, highlighting specific design considerations tailored to the integration of embodied AI in robotics.
We show on Figure~\ref{fig:rosllm-framework} an illustrative summary of the proposed framework.


\subsection{Atomic action library}
\label{sec:atomic-action-lib}

In both industrial and research settings, the development of robotic tasks often involves breaking down complex behaviors into simpleratomic actions. For instance, it is possible to complete a longer-horizon task such as pick-and-place by breaking it down into (i) acquire target, (ii) reach, (iii) grasp, (iv) acquire new target, (v) reach, and (vi) place.
While atomic actions commonly include physical movements, such as reaching or grasping, they can also include perceptual tasks like object detection or localization. 
Over time, a repertoire of tasks expands and the resulting library of atomic actions can facilitate the reusability of robotic behaviors.
While an initial set of these actions can be initially provided, it is unlikely to cover all potential tasks.
Therefore, our system includes a facility allowing non-expert users to add extra atomic actions to the library via imitation learning (see Section \ref{sec:imitation-learning} for further details).



In our framework, we integrate the concept of atomic actions with ROS and implement each atomic action as either a ROS action or a ROS service.
Additionally, for each action or service within the library, we provide a textual description to convey its intended functionality and usage.
This information is stored in a \texttt{JSON} file with the following fields: 
\texttt{`name`} of the ROS action/service, 
\texttt{`type`} that specifies if the atomic action is a ROS action or ROS service,
\texttt{`description'} of the atomic action and its input/output.  
We also designed tools, exposed in the ROS environment, to readily retrieve the action library description in the form of a readable string that can be exposed to the LLM through prompting.
These atomic actions can then be combined and orchestrated by the LLM agent to execute complex behaviors.

In addition, we provide functionality for the atomic action library, which is described by a code API that interfaces with the system.
In this case, the \texttt{JSON} field \texttt{`type'} is given the value \texttt{`code'}, and the behavior output from the language model can be an executable Python script.
Note that in this case, it is possible to have an action library combining ROS actions, ROS services, and a code API. 
However, the output of the language model is restricted to executable Python scripts.

\subsection{Environment observation}

In our framework, we operate under the assumption that the robot has access to various sensors capable of observing changes in the environment. 
These sensors provide valuable information about the state of the surroundings, which is essential for the robot to make informed decisions and adapt its behavior accordingly.
Given that our agent is a language model, we require environmental observations to be represented in textual form to serve as input to the model.

To facilitate this process, we implement a ROS package called the \texttt{`observation\_manager'}. 
This package implements tools that are responsible for querying multiple sensors and gathering textual observations of the environment. 
Upon initialization, a ROS node is configured with a list of services to call whenever an observation is needed. 
This modular design allows a user of our framework to customize the set of sensors they want to use based on their specific application requirements.

To set up our framework, we require users to implement these services that map observations from sensors to textual representations. 
For instance, a service may output descriptions such as ``the gripper is open'' or ``the blue box is detected in the camera view''. 
By standardizing the format of environmental observations as text, we promote modularity and interoperability within the framework.

\subsection{Human non-expert interface}

We provide a chat-based interface to our framework to allow easy adoption from non-expert human users.
Each environment step is executed after the human feedback is received from the interface, and then once the execution is over on the system (ending  either with a success or a failure), we ask the human to input a new textual entry. 
At the beginning, we let  the system interpret the first human input as the task description.
The task description should outline the goal or objective to assign to the robot, providing context for the subsequent actions to generate.
Thereafter, the system treats human input prompts as a feedback, which
 may contain suggestions for corrective behavior or suggestions for alternative approaches for the robot to complete the task.

Another potential interaction mode could be via speech, which would have the potential to be even more intuitive for non-experts.
We actually plan to implement a microphone into our setup and use an off-the-shelf audio-to-text package for parsing the input.
This functionality will be incorporated into our main code-base in the future.


\subsection{Prompt generation}

The prompt provided to the language model serves as input to generate behavior representations that can be executed on the system. 
At each environment step $\tau$, the prompt is updated, ensuring that the language model receives the latest information necessary for decision-making. 
We show in the central part of  Figure~\ref{fig:rosllm-framework} the different elements that we expect in a prompt to  shape the behavior of the system.

The prompt includes a task description that is provided by the user, as described in the previous sub-section.
After the first environment step $\tau$, the non-expert provides feedback that the system uses to correct its behavior.
A description of the atomic action library is also included to provide context on the admissible behaviors of the system, as described in Section~\ref{sec:atomic-action-lib}.
Moreover, the prompt contains an observation of the environment that is collected by mapping several sensor readings to text.
Several well-known prompt engineering strategies are utilized to aid the language model construct a behavior for the system, namely chain-of-thought and few-shot prompting. 
These additional portions of the prompt are assumed to be given as part of all observations $o$.
Finally, some additional notes are written in the prompt, such as how the language model should format the behavior output (e.g. Python or XML).

Overall, the prompt generation process gathers information from both the ROS environment and the human interface, ensuring that the language model receives comprehensive input to guide its decision-making process. 
Once the prompt is constructed, it is then passed to the language model.
We consider the output of the language model to represent the desired behavior for the system.
We describe next the different formatting options for the  output of the language model.

\subsection{Behavior representation}

We call a \textit{behavior} the combination of atomic actions that is extracted from the textual output of the language model.
To represent a behavior, the LLM generates either a Python or XML code.
When Python format is used,  a Python terminal  exposed to the ROS environment executes the LLM output.
In the case of XML, the LLM response is interpreted as a behavior tree\footnote{The behavior tree interface has recently been added to the code base. We plan in the future to run experiments and report the results in an updated version of this article.}.
We use regular expressions to easily identify parse LLM output that should encapsulate the code into
\texttt{{`}{`}{`}python...{'}{'}{'}},
\texttt{{`}{`}{`}json...{'}{'}{'}}, or
\texttt{{`}{`}{`}xml...{'}{'}{'}}.

\paragraph{Python output} We expect the use of Python code when the action library is a set of Python function that exposes the various functionality of the system.
The library can also contain ROS actions and ROS services that can be interfaced with the script.

\paragraph{JSON output} When the JSON format is used, a behavior representation called an action sequence is used. 
In this case, the specified actions in the sequence are executed one after the other, and we expect each action to be a ROS service of the type \texttt{rosllm\_srvs/AtomicAction}.
This service returns a string called \texttt{output} and takes as input one string argument called \texttt{input}, which takes input per action, and 
another string argument \texttt{prev\_output}, which is the output from the previous action.

\paragraph{XML output} A behavior tree, represented by XML code, is a hierarchical model that describes the behavior of autonomous agents or robots. 
It consists of nodes that define specific actions or conditions, and it is organized in a tree-like structure.
At the root of the tree, the behavior selector node determines the order in which to evaluate and execute child nodes. 
These child nodes can include sequences of nodes, which execute their child nodes sequentially until one fails, or parallel nodes, which execute their child nodes simultaneously. 
Other types of nodes include conditional nodes, action nodes, and decorator nodes, each serving distinct roles in controlling the agent's behavior.
In this case, the language model is tasked with producing the XML code defining a behavior tree.

\subsection{Updating the atomic action library via imitation learning}\label{sec:imitation-learning}

In our framework, the atomic action library may not encompass all the tasks that non-expert users want to realize. 
To address this limitation, we provide a mechanism for non-experts to update the action library using imitation learning techniques. 
This section details our approach to integrating imitation learning into our framework and describes the process of updating the action library through kinesthetic teaching or teleoperation.

As a first step, we integrate to our framework a connection to dynamic movement primitives (DMP), which is a widely-used technique requiring a minimal number of demonstrations to learn smooth representations of tasks.
However, our framework remains flexible and can accommodate other imitation learning pipelines, including behavior cloning with deep learning, depending on the specific requirements of the task.

When a non-expert provides a demonstration of a task along with a textual description, our framework automatically translates this information into a ROS service and adds it to the atomic action library. 
The demonstration is captured using kinesthetic teaching or teleoperation, allowing the non-expert to guide the robot through the desired behavior.
Subsequently, the learned task is represented as an atomic action within the library, while the accompanying textual description is included in the action library descriptions.
This process enables the framework to expand its repertoire of available actions based on user demonstrations and textual annotations, thereby enhancing its adaptability and versatility in addressing a wide range of tasks.

By empowering non-experts to contribute to the expansion of the action library through imitation learning, our framework promotes collaborative human-robot interaction. which could accelerate the development of robot capabilities tailored to specific users' needs.

\section{Experiments}

In this section, we present a series of experiments to evaluate the performance and capabilities of our framework for intuitive robot programming.
These experiments encompass diverse scenarios, ranging from long-horizon tasks to dynamic environment adaptations and intercontinental supervisory control.
Through these experiments, we assess the system's ability to comprehend natural language prompts, to generate accurate action sequences, to adapt to changing environments, and to collaborate effectively with human operators. 
The experiments showcase our framework's robustness, scalability, and versatility, enabling non-experts to program robots effortlessly.
Thus, we are advancing the state-of-the-art in embodied AI and human-robot collaboration.

\subsection{Experiments setup}

\begin{figure}[t]
    \centering
    \includegraphics[width=\textwidth]{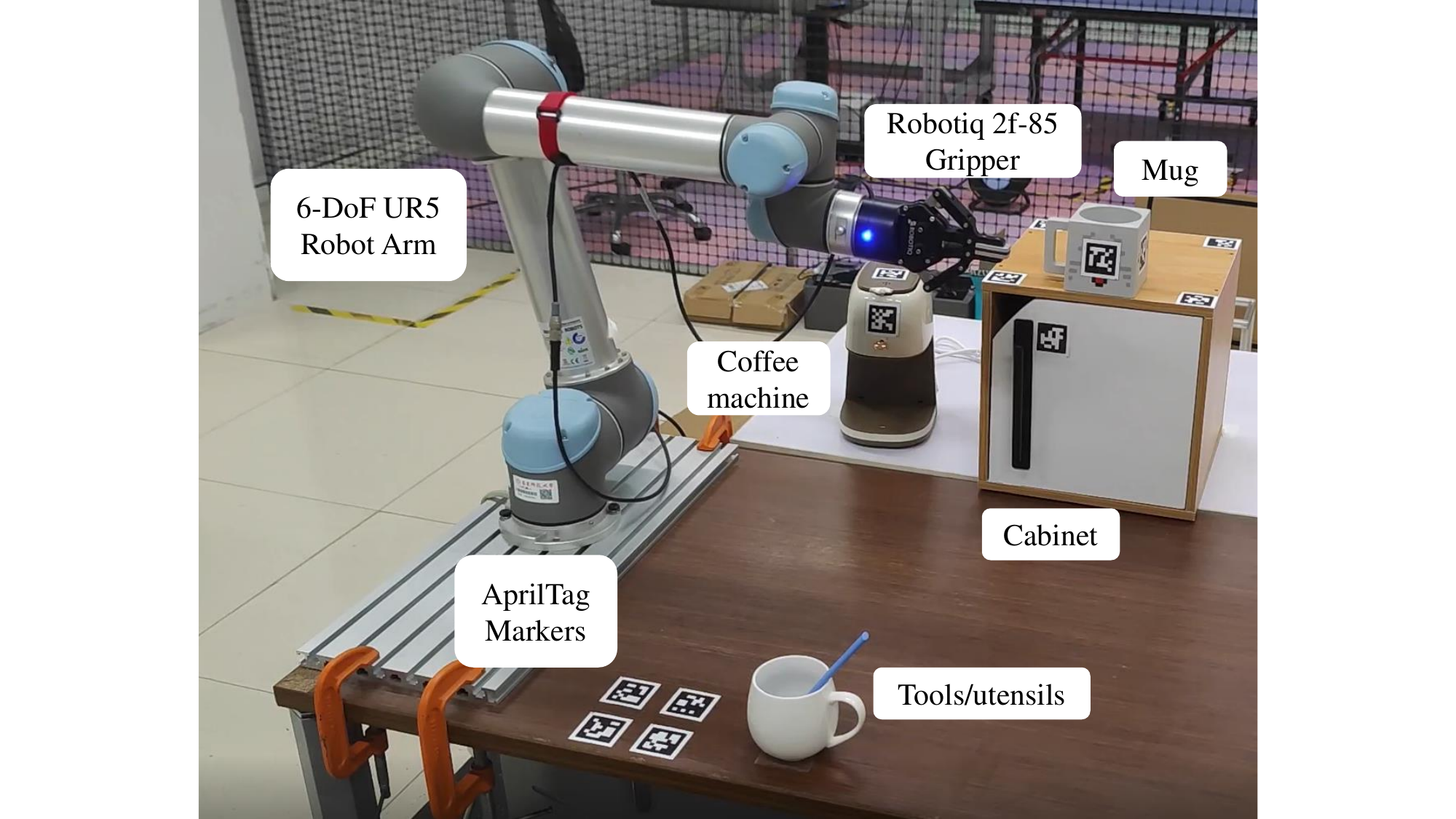}
    \caption{Real-world laboratory setup used in our experiments.}
    \label{fig:real-world-setup}
\end{figure}

Our real-world robot setup, shown in Figure~\ref{fig:real-world-setup}, comprises a Universal Robots UR5 arm equipped with a Robotiq 2-finger gripper (2F-85) attached at the end-effector. 
Situated within a kitchen-like environment, our setup includes various objects such as boxes, a bowl, utensils (e.g., a spoon), a sink, and a cabinet with a door. 
We strategically place these objects around the robot within its workspace to simulate real-world scenarios.
The robot is controlled using the ROS Noetic, utilizing standard packages tailored for UR/Robotiq robots. 
Our control system runs on a laptop powered by Ubuntu 20.04, featuring a 16-core i9 Intel CPU, which ensures robust performance and efficient task execution.

For language processing and task execution, we utilize the Deepseek 7B Coder, an open-source LLM deployed on a virtual LLM (vLLM) server located in Europe.
Communication between the robot setup in Asia and the vLLM server is established via the internet, with requests transmitted over HTTP protocol.

To facilitate seamless interaction and task execution, we have developed a chat interface integrated with an AI agent, a novel framework for integrating and learning structured reasoning into AI agents’ policies. 
This chat interface enables intuitive communication with the robot system, allowing users to articulate task requirements and provide feedback effortlessly.
Furthermore, we incorporate Whisper~\cite{radford2023robust}, a tool that translates audio into text, offering an alternative interface through spoken language.

For perception and state estimation, we use an Intel Realsense RGBD camera positioned at a fixed location facing the robot and scene.
Despite the camera's depth-sensing capabilities, we use only RGB data channels for computational efficiency to simplify our implementation. 
In the environment, we add fiducial marker tags and utilize capabilities from the AprilTag library~\cite{Olson11} for sensing their poses with respect to the camera frame.
To estimate the transformation between the camera the and robot base frame, we use standard eye-to-base calibration packages from the Moveit library.

In summary, our real-world robot setup leverages state-of-the-art hardware and software components, coupled with advanced language processing capabilities, to enable intuitive interactions and successful task execution within a dynamic and realistic environment.

\subsection{Long-horizon tasks}

\begin{figure}[ht]
\centering
\begin{tikzpicture}[
    node distance=1.5cm and 1.0cm, 
    auto,
    every node/.style={align=center, ellipse, draw=black, minimum width=2.3cm, minimum height=1cm, text width=2cm},
    arrow/.style={-Stealth, thick}
  ]

  \node (step1) {Picking up the mug};
  \node[right=of step1] (step2) {Placing the mug\\ in the machine};
  \node[right=of step2] (step3) {Opening the\\ machine cover};

  \node[below=of step3] (step4) {Opening the\\ cabinet door};
  \node[left=of step4] (step5) {Picking up\\ and place \\ the bowl};
  \node[left=of step5] (step6) {Picking up\\ the spoon};

  \node[below=of step6] (step7) {Scooping coffee\\ from the bowl};
  \node[right=of step7] (step8) {Placing coffee\\ in the machine};
  \node[right=of step8] (step9) {Closing the\\ machine cover};

  \node[below=of step9] (step10) {Returning the bowl\\ to the cabinet};
  \node[left=of step10] (step11) {Closing the\\ cabinet door};
  \node[left=of step11] (step12) {Switching on\\ the coffee machine};

  \draw[arrow] (step1) -- (step2);
  \draw[arrow] (step2) -- (step3);
  
  \draw[arrow] (step3) -- (step4); 
  \draw[arrow] (step4) -- (step5);
  \draw[arrow] (step5) -- (step6);

  \draw[arrow] (step6) -- (step7); 
  \draw[arrow] (step7) -- (step8);
  \draw[arrow] (step8) -- (step9);

  \draw[arrow] (step9) -- (step10); 
  \draw[arrow] (step10) -- (step11);
  \draw[arrow] (step11) -- (step12);

\end{tikzpicture}
\caption{Detailed steps in the coffee-making process arranged in a modified Z-shaped flow across four rows.}
\label{fig:coffee-zshape}
\end{figure}
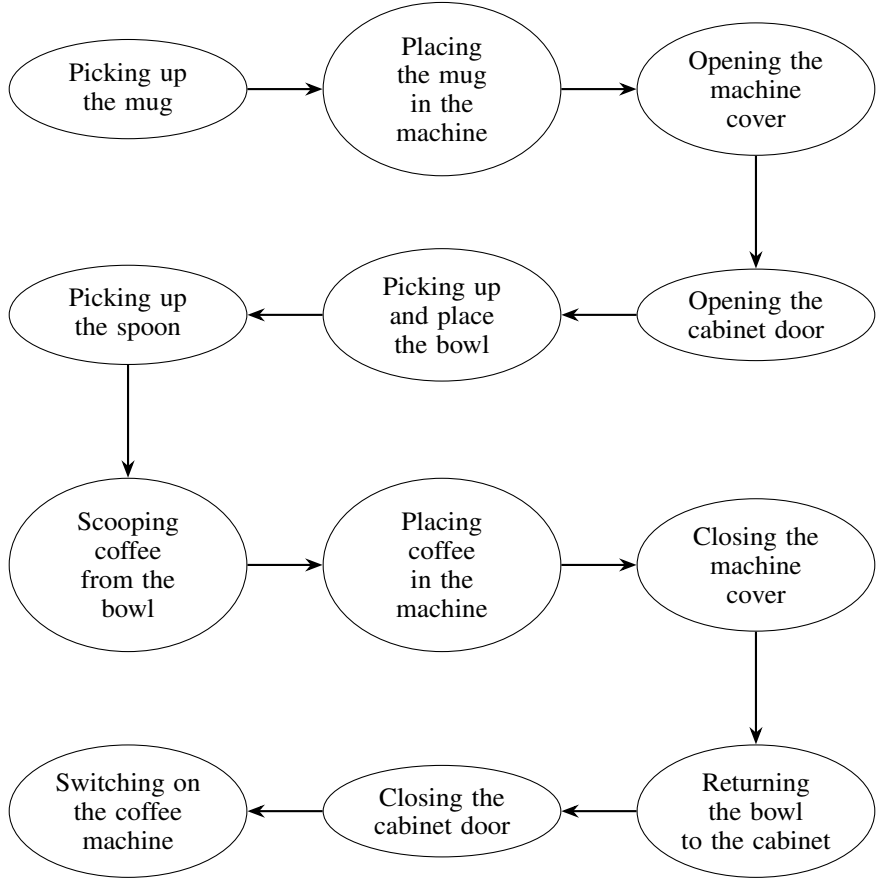

\begin{figure}[ht]
\centering

\begin{subfigure}{0.16\textwidth}
  \centering
  \includegraphics[width=\linewidth]{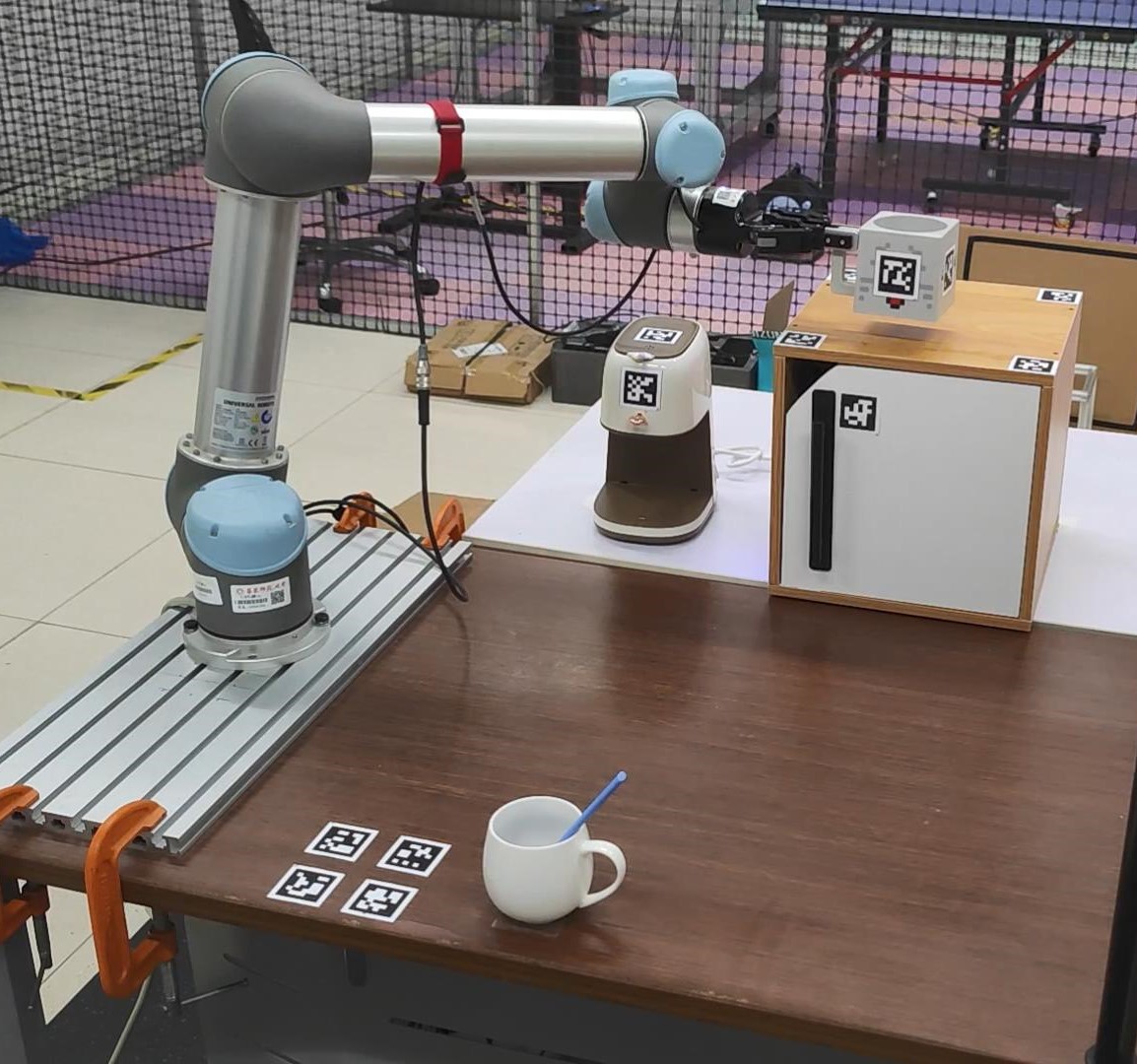}
  \caption{} 
  \label{fig:coffee1}
\end{subfigure}%
\hfill
\begin{subfigure}{0.16\textwidth}
  \centering
  \includegraphics[width=\linewidth]{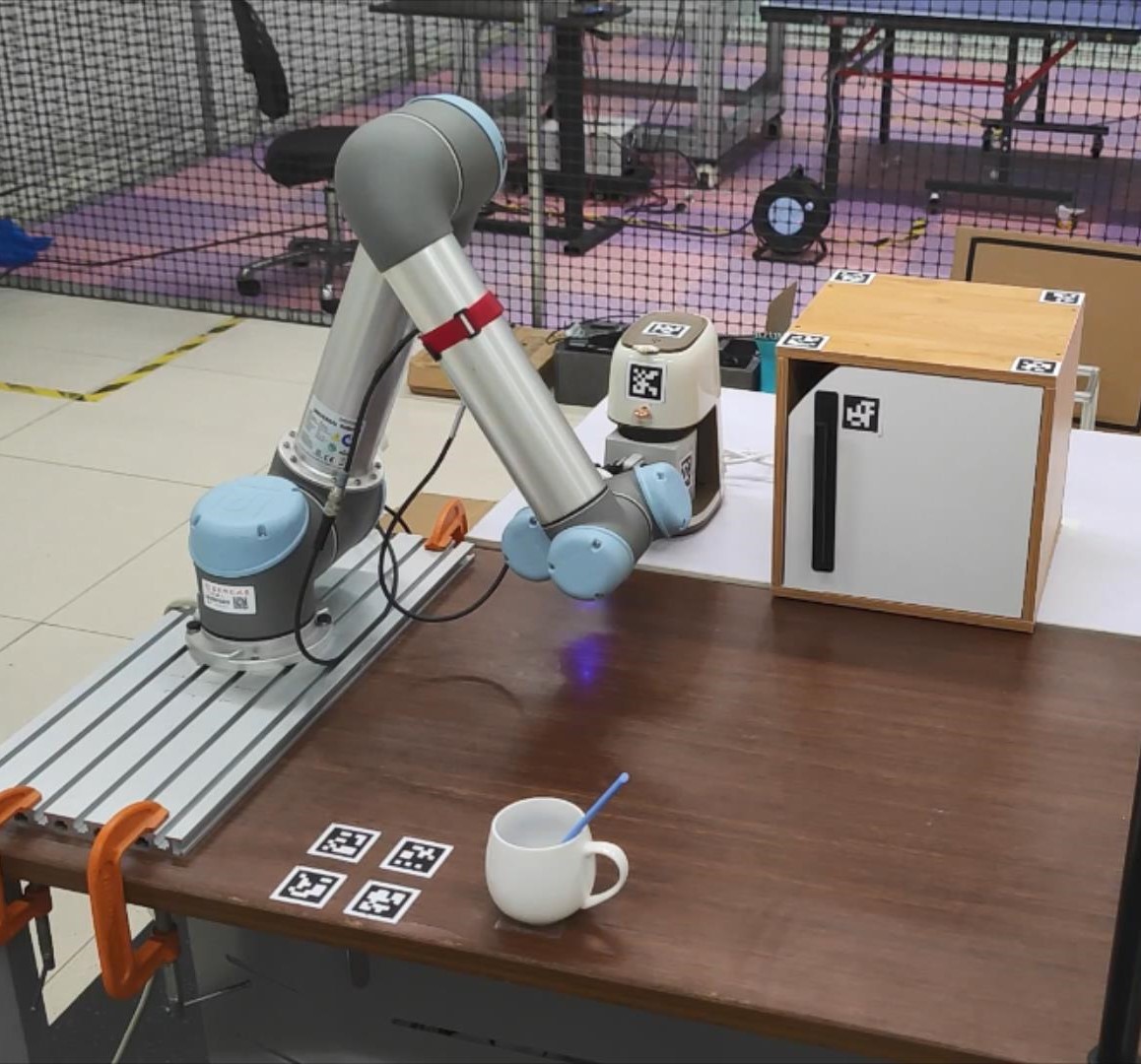}
  \caption{} 
  \label{fig:coffee2}
\end{subfigure}%
\hfill
\begin{subfigure}{0.16\textwidth}
  \centering
  \includegraphics[width=\linewidth]{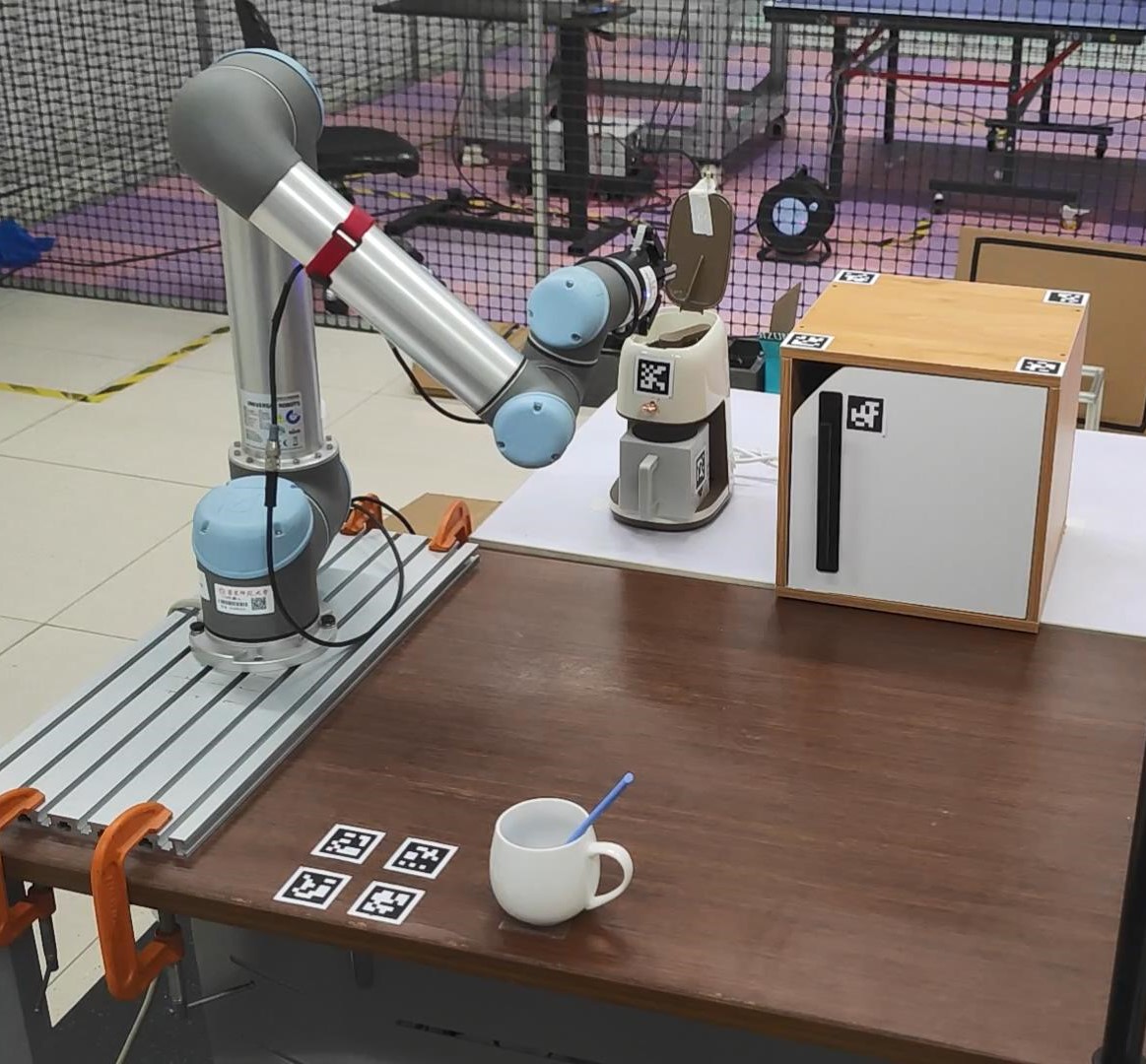}
  \caption{} 
  \label{fig:coffee3}
\end{subfigure}%
\hfill
\begin{subfigure}{0.16\textwidth}
  \centering
  \includegraphics[width=\linewidth]{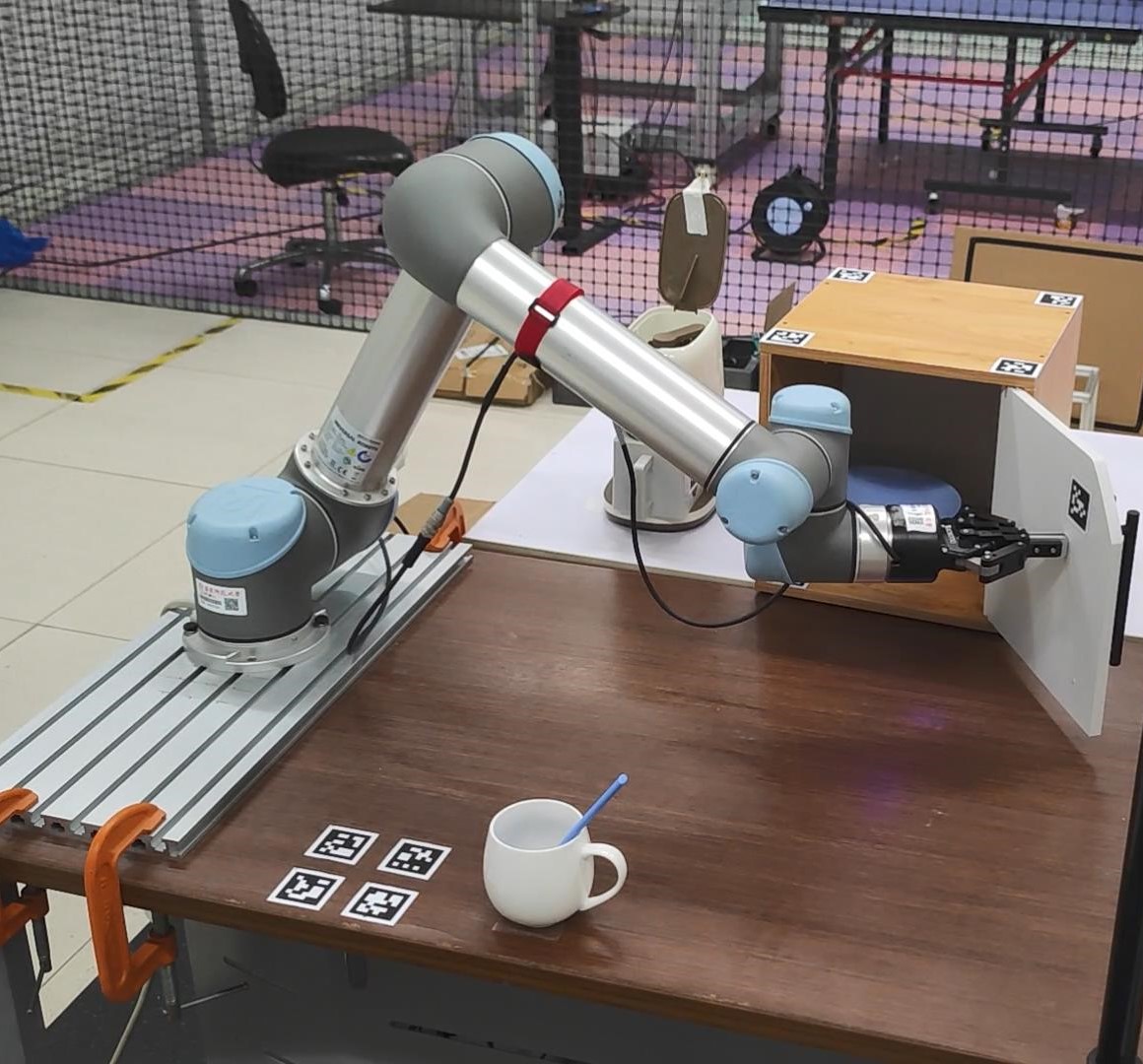}
  \caption{} 
  \label{fig:coffee4}
\end{subfigure}%
\hfill
\begin{subfigure}{0.16\textwidth}
  \centering
  \includegraphics[width=\linewidth]{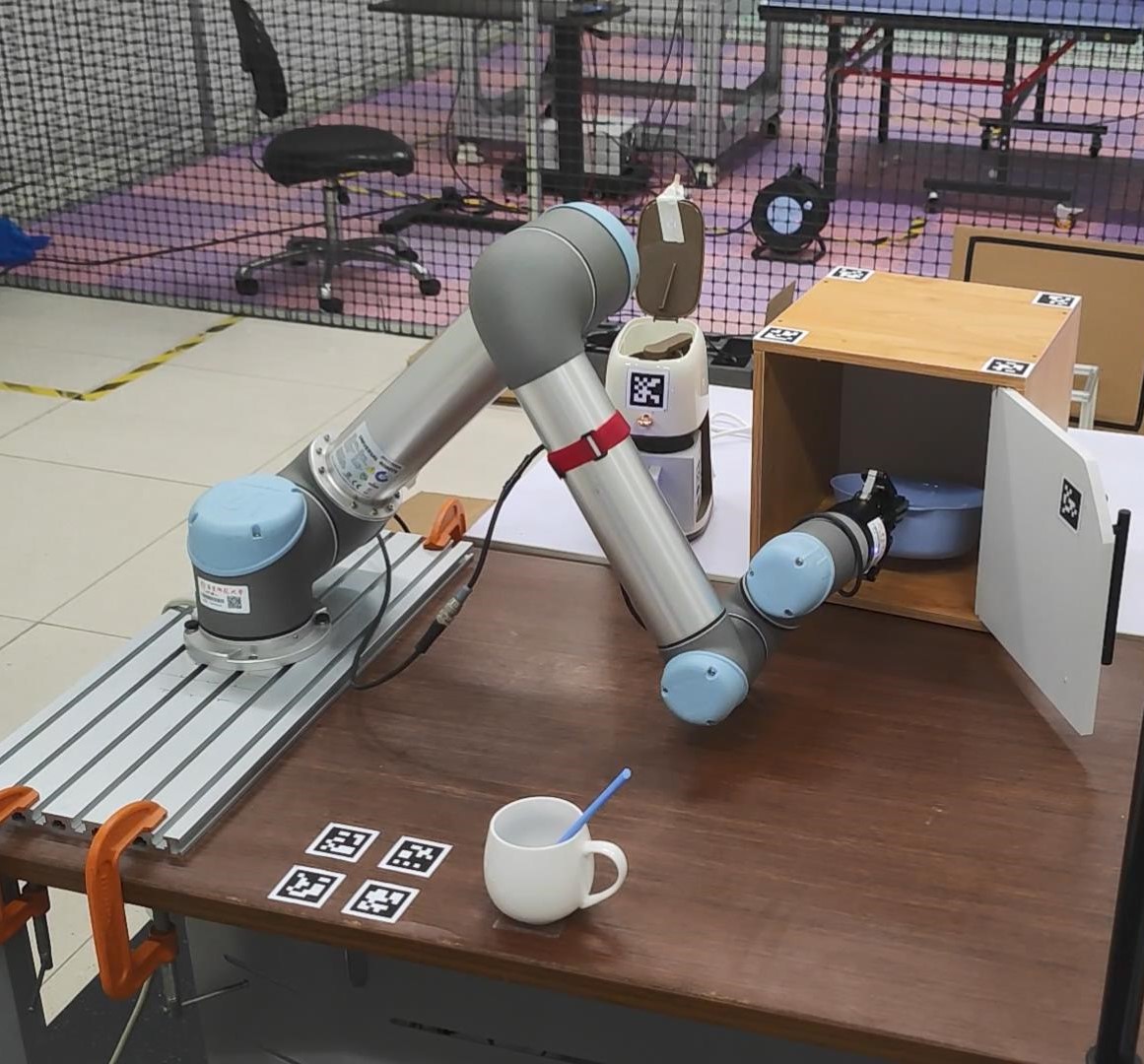}
  \caption{} 
  \label{fig:coffee5}
\end{subfigure}%
\hfill
\begin{subfigure}{0.16\textwidth}
  \centering
  \includegraphics[width=\linewidth]{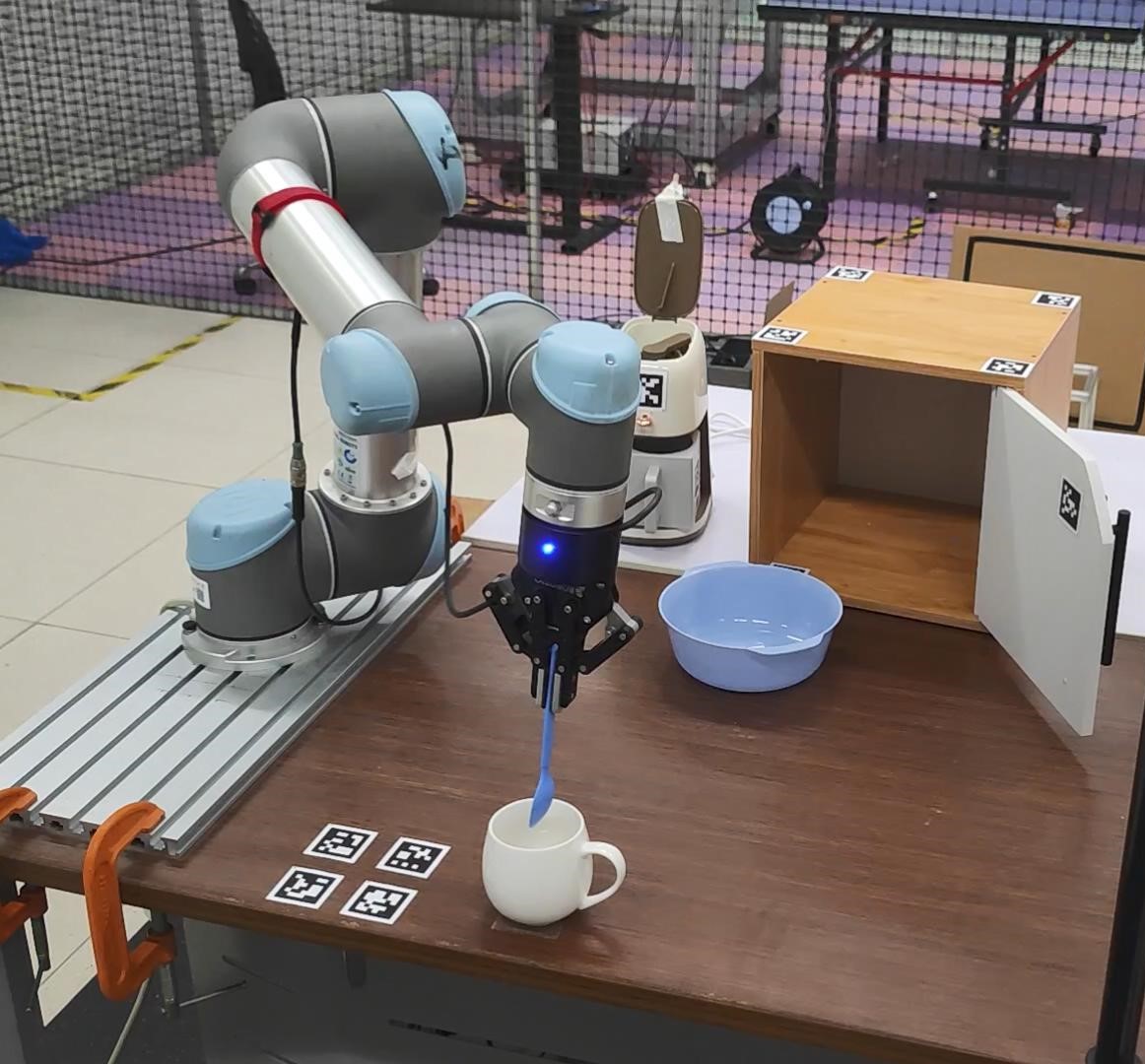}
  \caption{} 
  \label{fig:coffee6}
\end{subfigure}

\begin{subfigure}{0.16\textwidth}
  \centering
  \includegraphics[width=\linewidth]{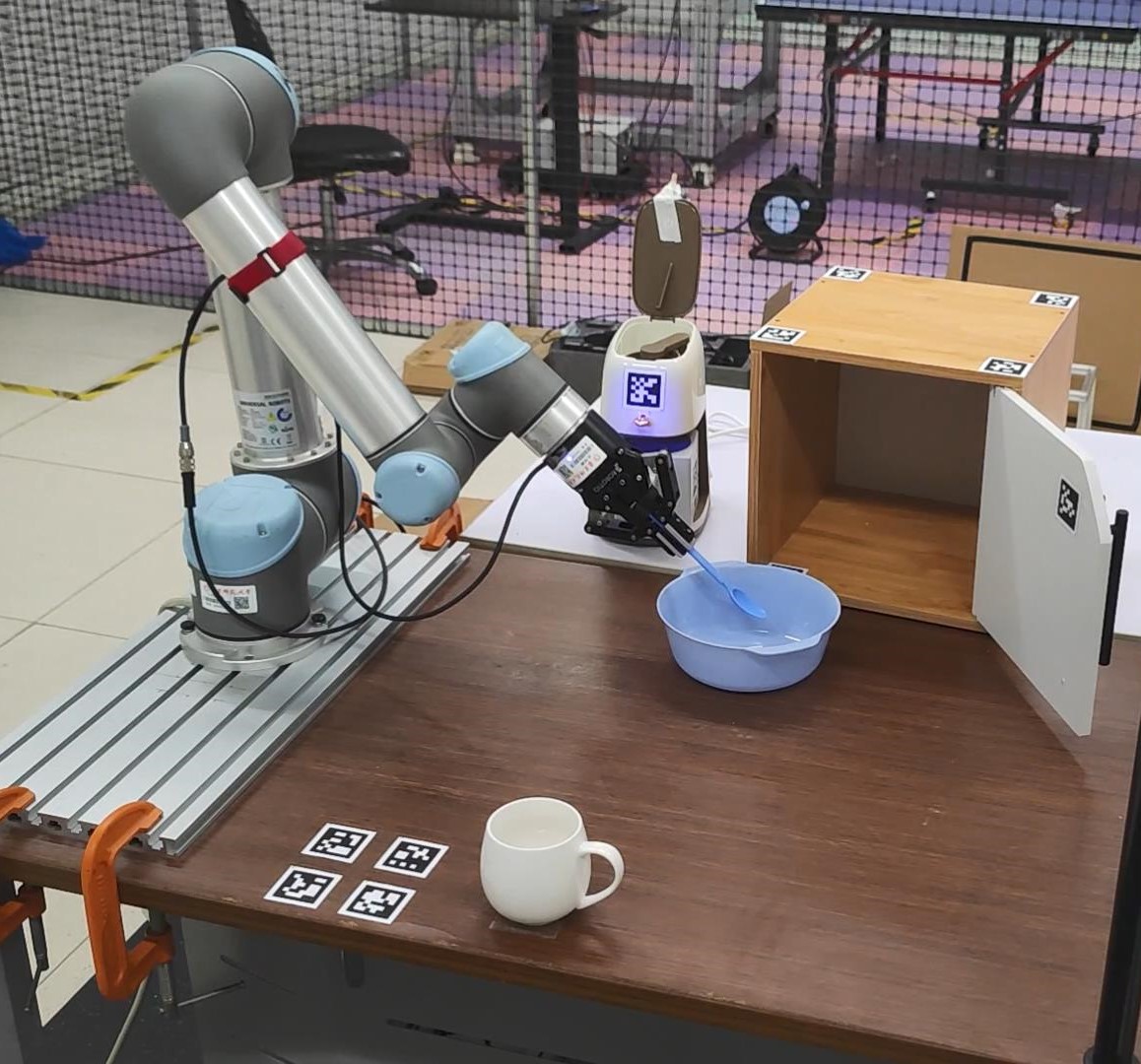}
  \caption{} 
  \label{fig:coffee7}
\end{subfigure}%
\hfill
\begin{subfigure}{0.16\textwidth}
  \centering
  \includegraphics[width=\linewidth]{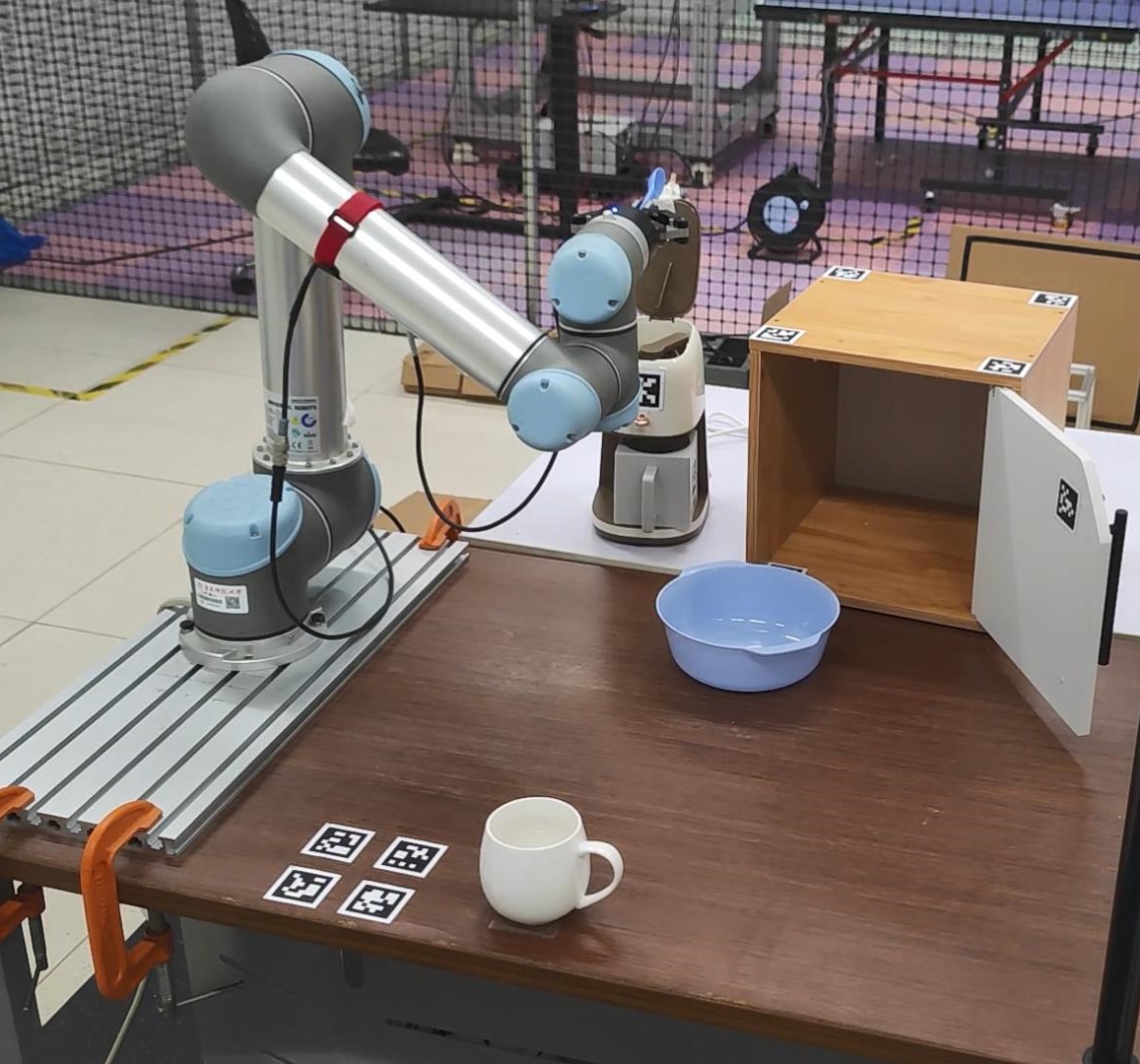}
  \caption{} 
  \label{fig:coffee8}
\end{subfigure}%
\hfill
\begin{subfigure}{0.16\textwidth}
  \centering
  \includegraphics[width=\linewidth]{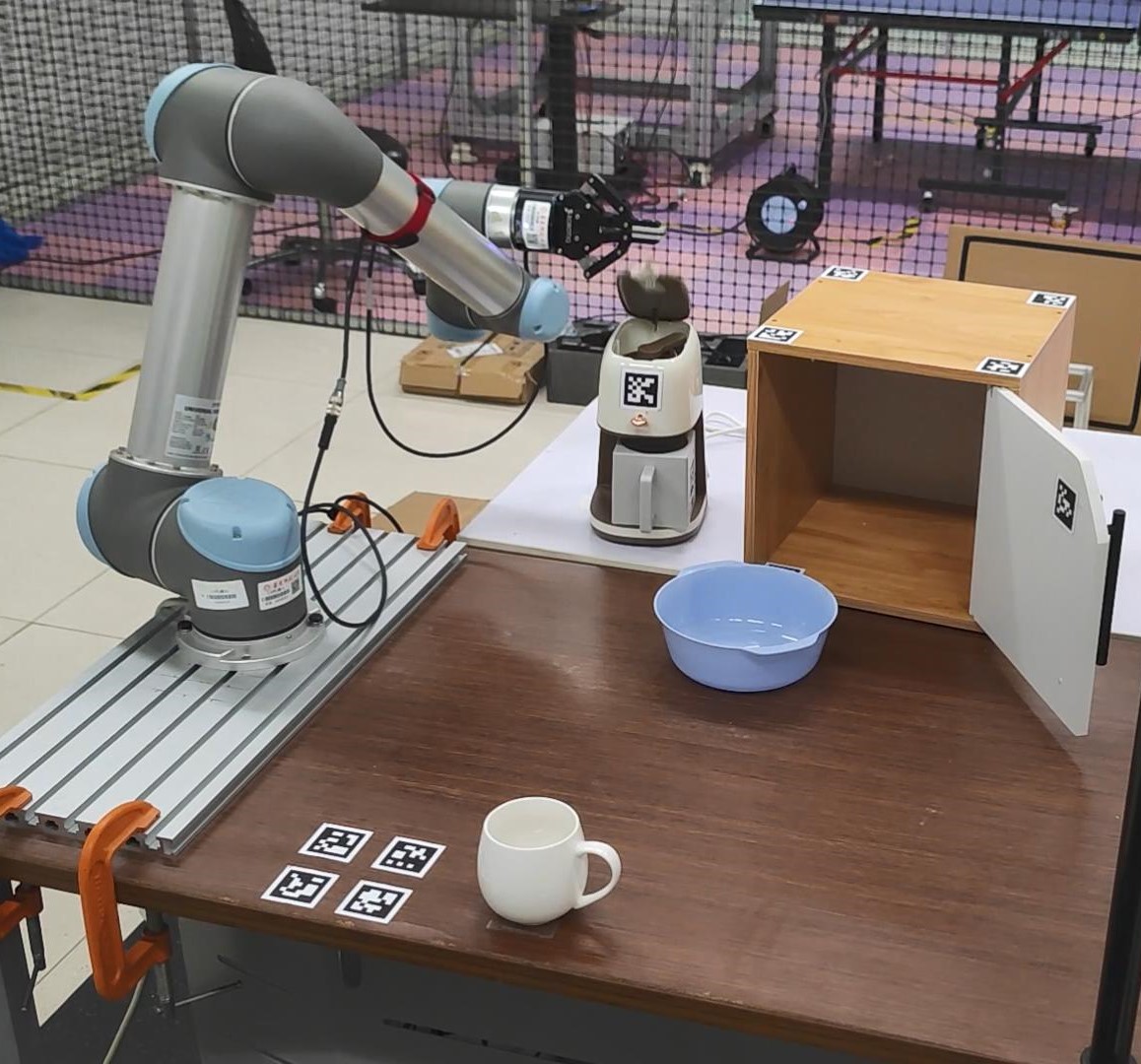}
  \caption{} 
  \label{fig:coffee9}
\end{subfigure}%
\hfill
\begin{subfigure}{0.16\textwidth}
  \centering
  \includegraphics[width=\linewidth]{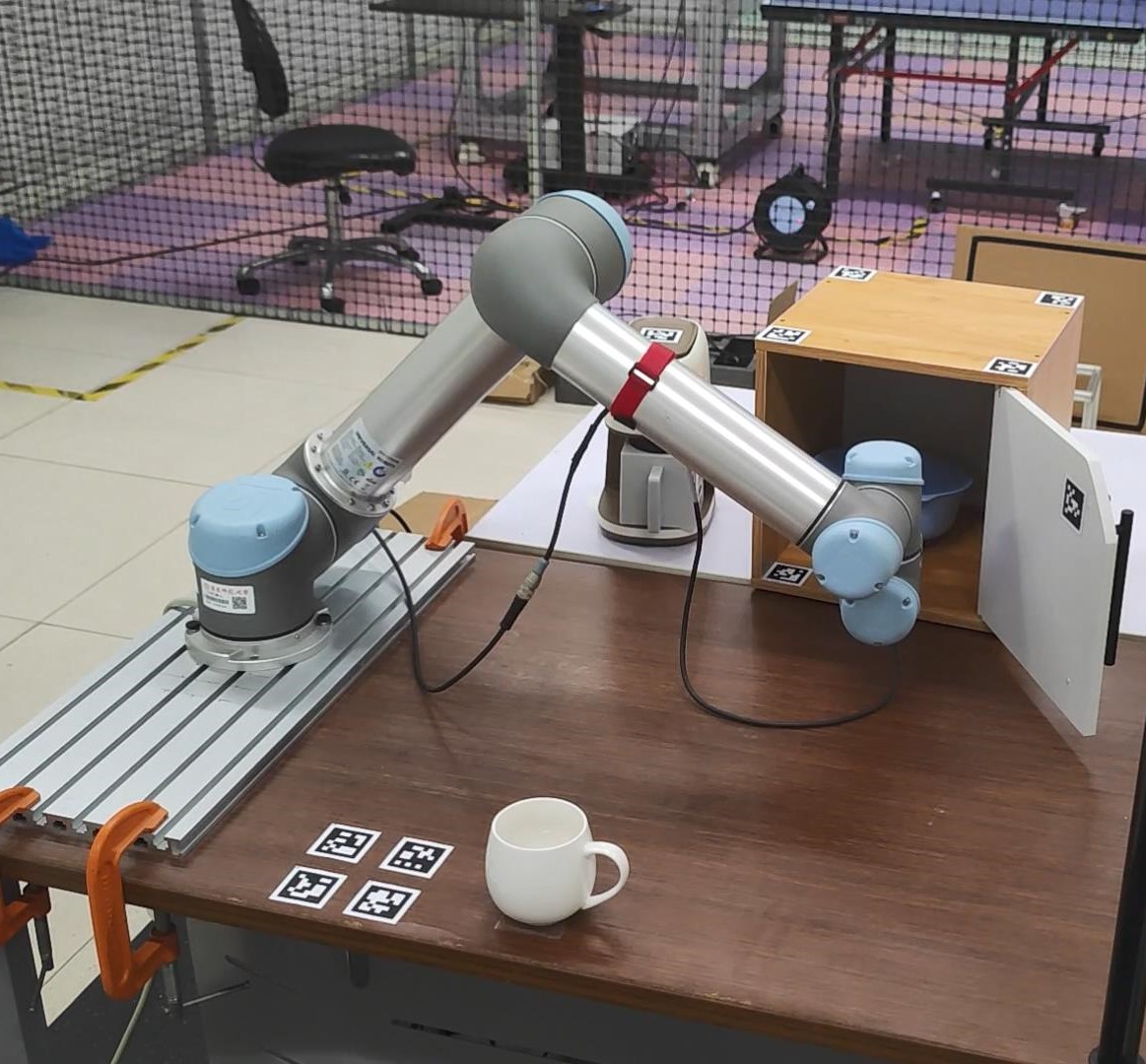}
  \caption{} 
  \label{fig:coffee10}
\end{subfigure}%
\hfill
\begin{subfigure}{0.16\textwidth}
  \centering
  \includegraphics[width=\linewidth]{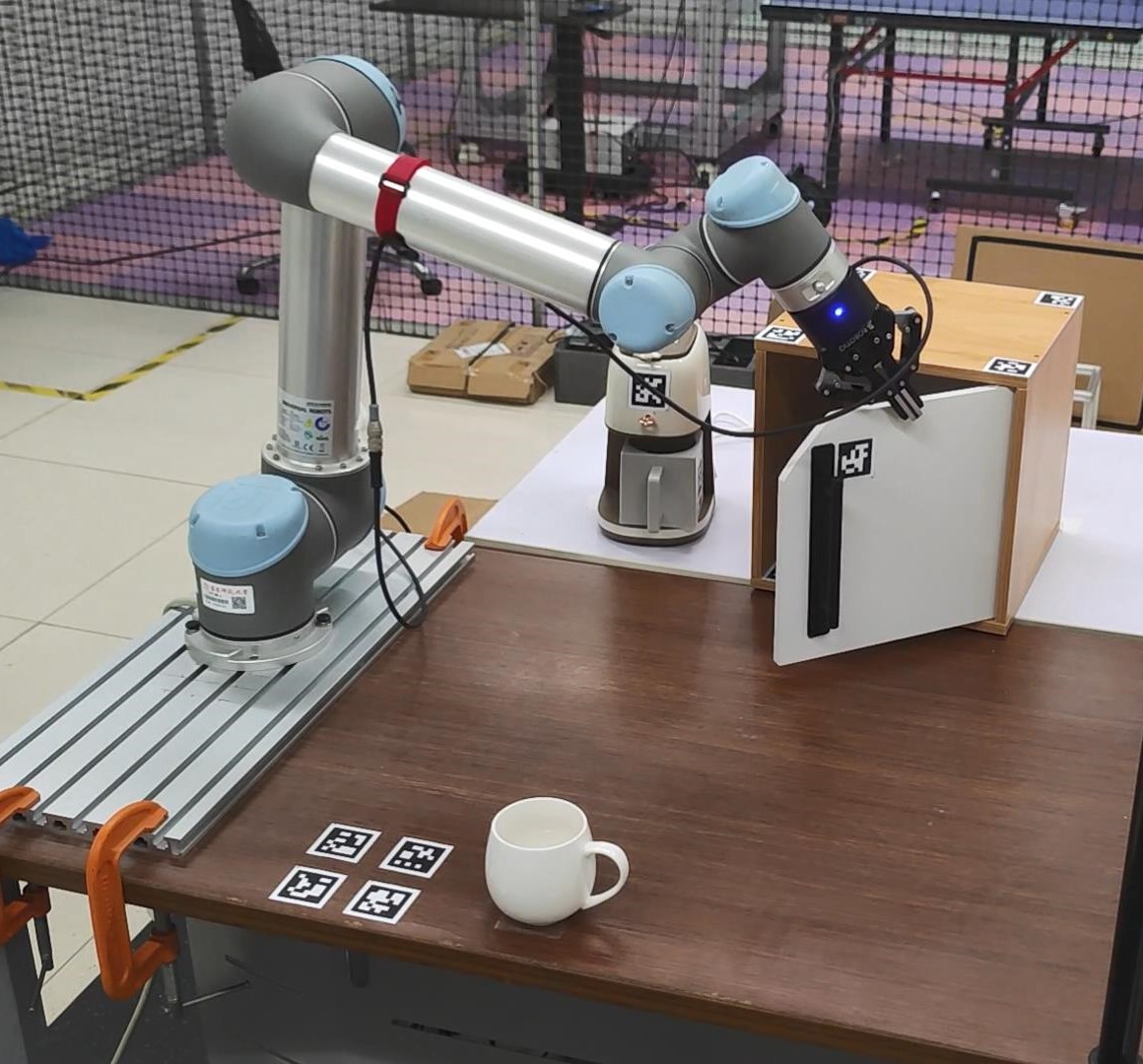}
  \caption{} 
  \label{fig:coffee11}
\end{subfigure}%
\hfill
\begin{subfigure}{0.16\textwidth}
  \centering
  \includegraphics[width=\linewidth]{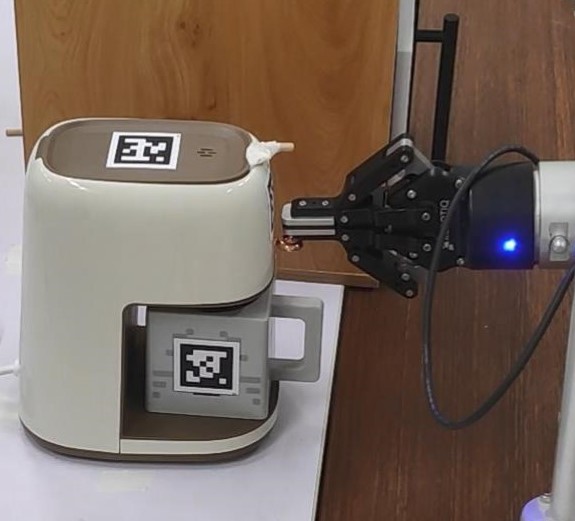}
  \caption{} 
  \label{fig:coffee12}
\end{subfigure}

\caption{Detailed steps in the coffee-making process are depicted across twelve images: (a) picking up the mug, (b) placing the mug in the coffee machine, (c) opening the coffee machine cover, (d) opening the cabinet door, (e) picking up the bowl, (f) picking up the spoon, (g) scooping coffee from the bowl, (h) placing the coffee in the machine, (i) closing the coffee machine cover, (j) returning the bowl to the cabinet, (k) closing the cabinet door, (l) switching on the coffee machine.}
\label{fig:coffee}
\end{figure}

The experiment on long-horizon tasks serves to evaluate the system's ability to comprehend and execute complex multistep tasks, exemplified by the request ``can you make me a coffee''. 
This experiment is essential as it represents a real-world scenario where robots need to perform sequential actions to accomplish a goal, requiring a combination of perception, manipulation, and decision-making skills.

In this experiment, we employ an open-source language model( Deepseek 7B Coder) to generate action sequences based on natural language prompts. 
The need for a detailed, natural language description of the task steps arises from the intricacies of the language model's comprehension capabilities. 

We provide the system with a comprehensive atomic action library encompassing a range of fundamental actions necessary for coffee preparation. 
These include reaching, opening/closing doors, picking up objects, placing objects, switching machines on/off, opening/closing the coffee machine cover, taking objects out of a cabinet, putting objects back in the cabinet, spooning coffee grounds from the bowl with a spoon, inserting the mug into the coffee machine, drop objects in the sink.

To execute the task of making coffee, the system first identifies and locates the necessary objects and equipment in the environment, such as the coffee machine, mug, coffee grounds, and spoon. 
Subsequently, it plans a sequence of actions to perform each step of the coffee-making process. 
This entails grasping and manipulating objects, navigating the environment, and interacting with the coffee machine and other appliances.

The system takes as input a task prompt that consists of a natural language description of the task steps, resembling a recipe for making coffee. 
This contextual information is crucial for guiding the system's understanding and ensuring the generation of accurate action sequences.

In Figure \ref{fig:coffee-zshape}, we show the key steps of the experiment, with corresponding photos in Figure \ref{fig:coffee}, illustrating the systematic execution of actions by the robot in the real-world. 
The successful outcome of the experiment demonstrates the system's capability to comprehend and execute long-horizon tasks accurately and efficiently.

Overall, this experiment underscores the effectiveness of our framework in enabling robots to tackle complex, multistep tasks through intuitive natural language prompts and comprehensive action libraries.

\subsection{Policy correction via human feedback}

\begin{figure}[ht]
\centering

\begin{subfigure}{0.19\textwidth}
  \centering
  \includegraphics[width=\linewidth]{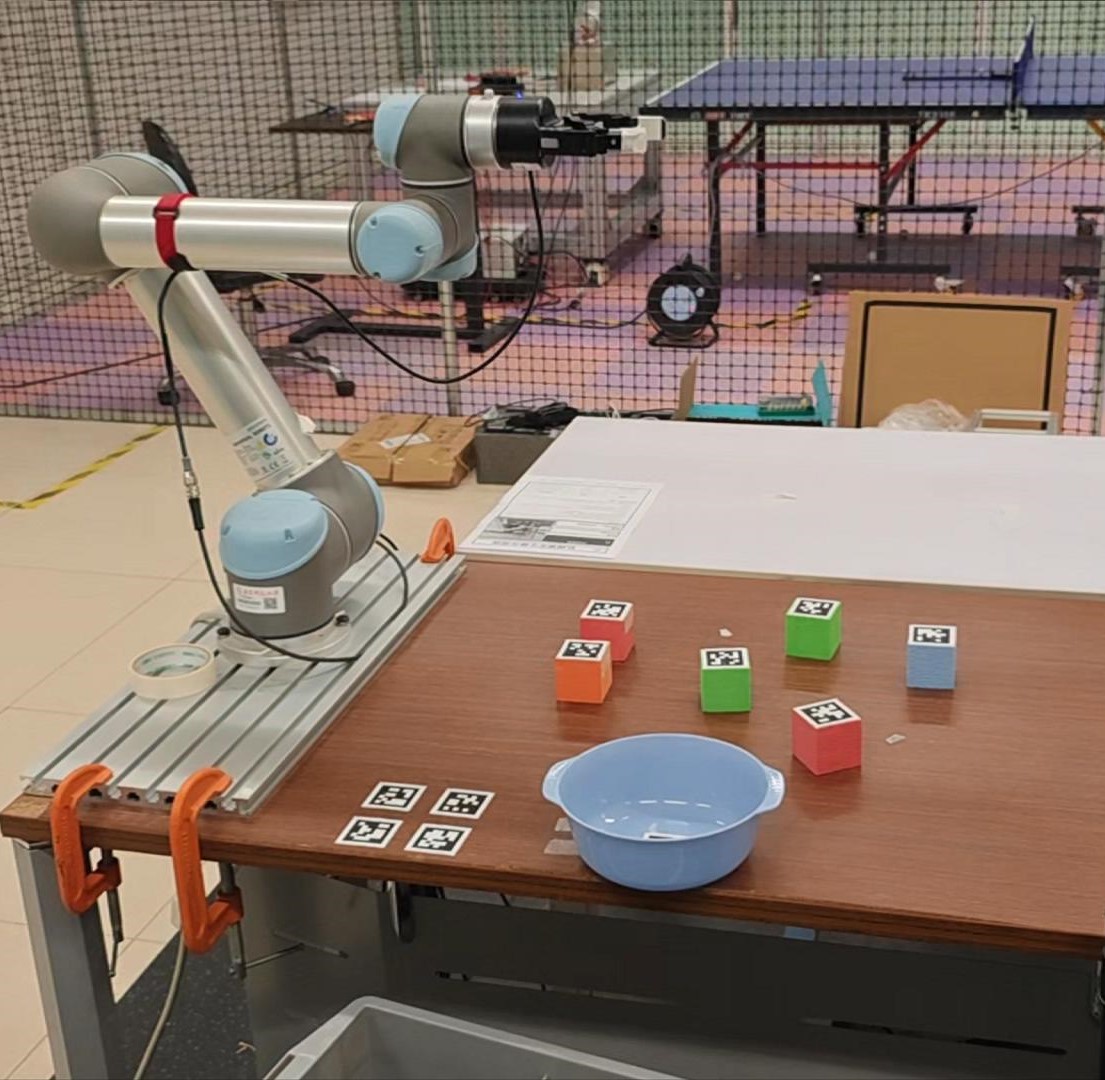}
  \caption{} 
  \label{fig:6cube1}
\end{subfigure}
\hfill
\begin{subfigure}{0.19\textwidth}
  \centering
  \includegraphics[width=\linewidth]{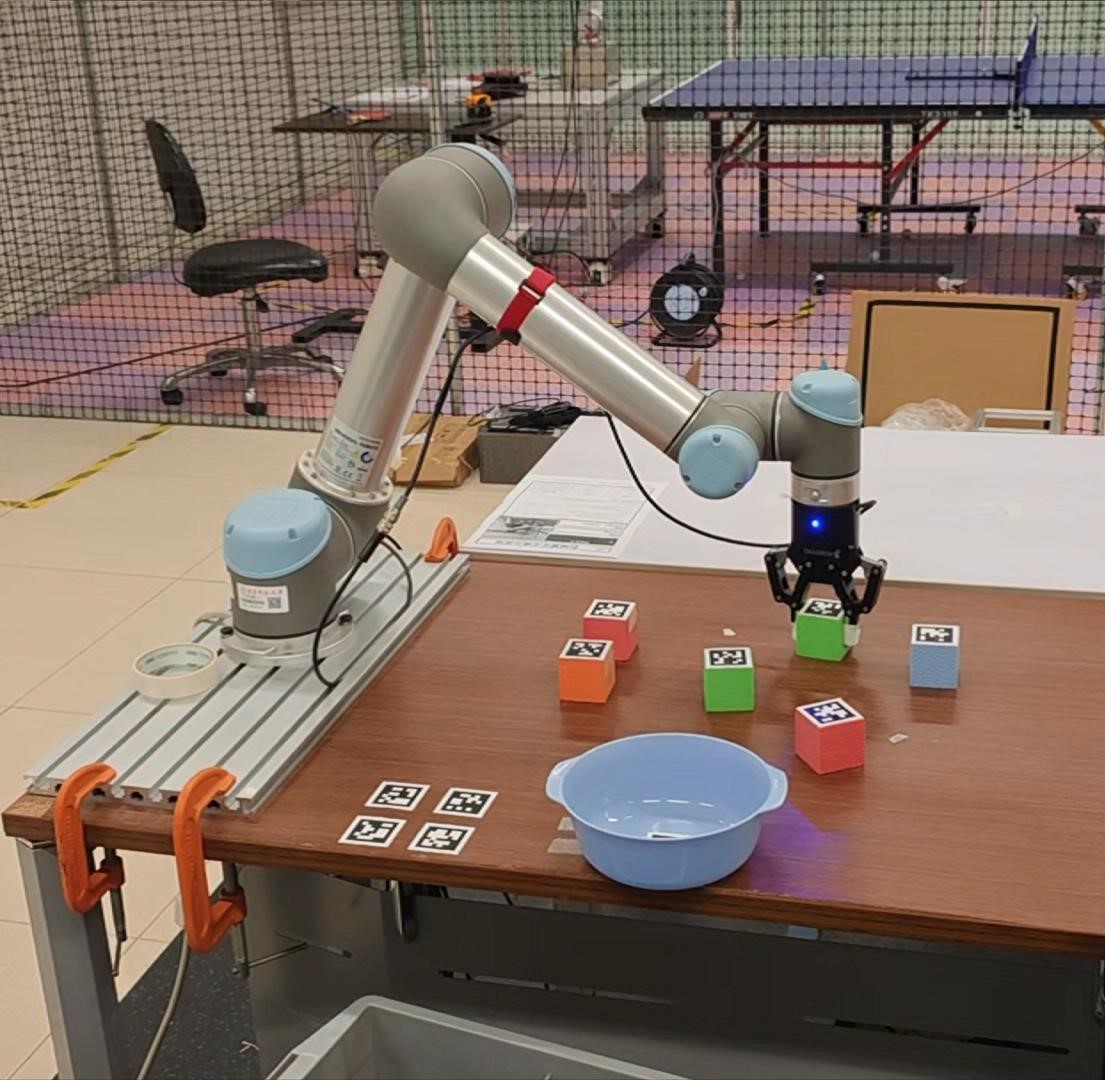}
  \caption{} 
  \label{fig:6cube2}
\end{subfigure}
\hfill
\begin{subfigure}{0.19\textwidth}
  \centering
  \includegraphics[width=\linewidth]{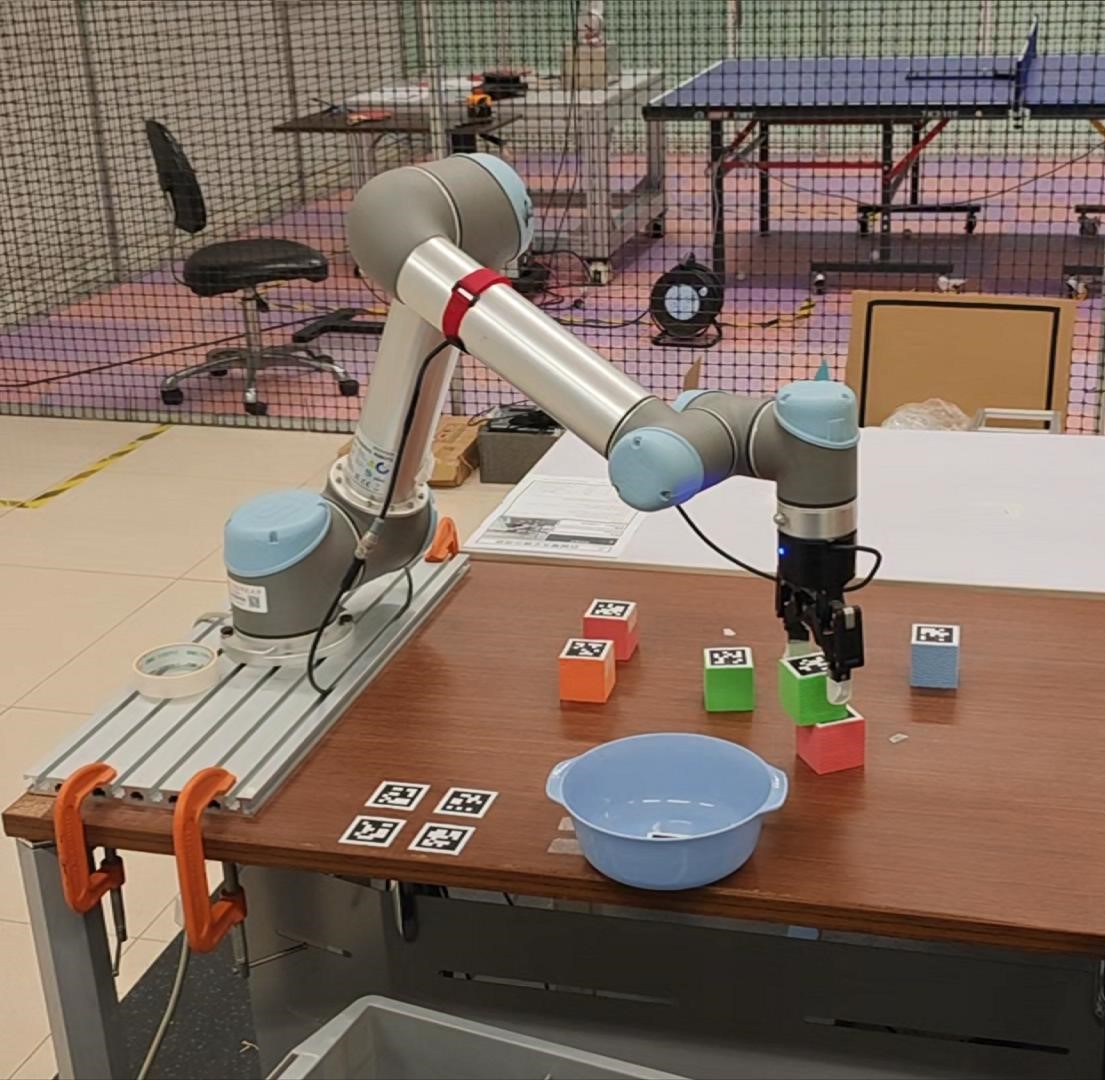}
  \caption{} 
  \label{fig:6cube3}
\end{subfigure}
\hfill
\begin{subfigure}{0.19\textwidth}
  \centering
  \includegraphics[width=\linewidth]{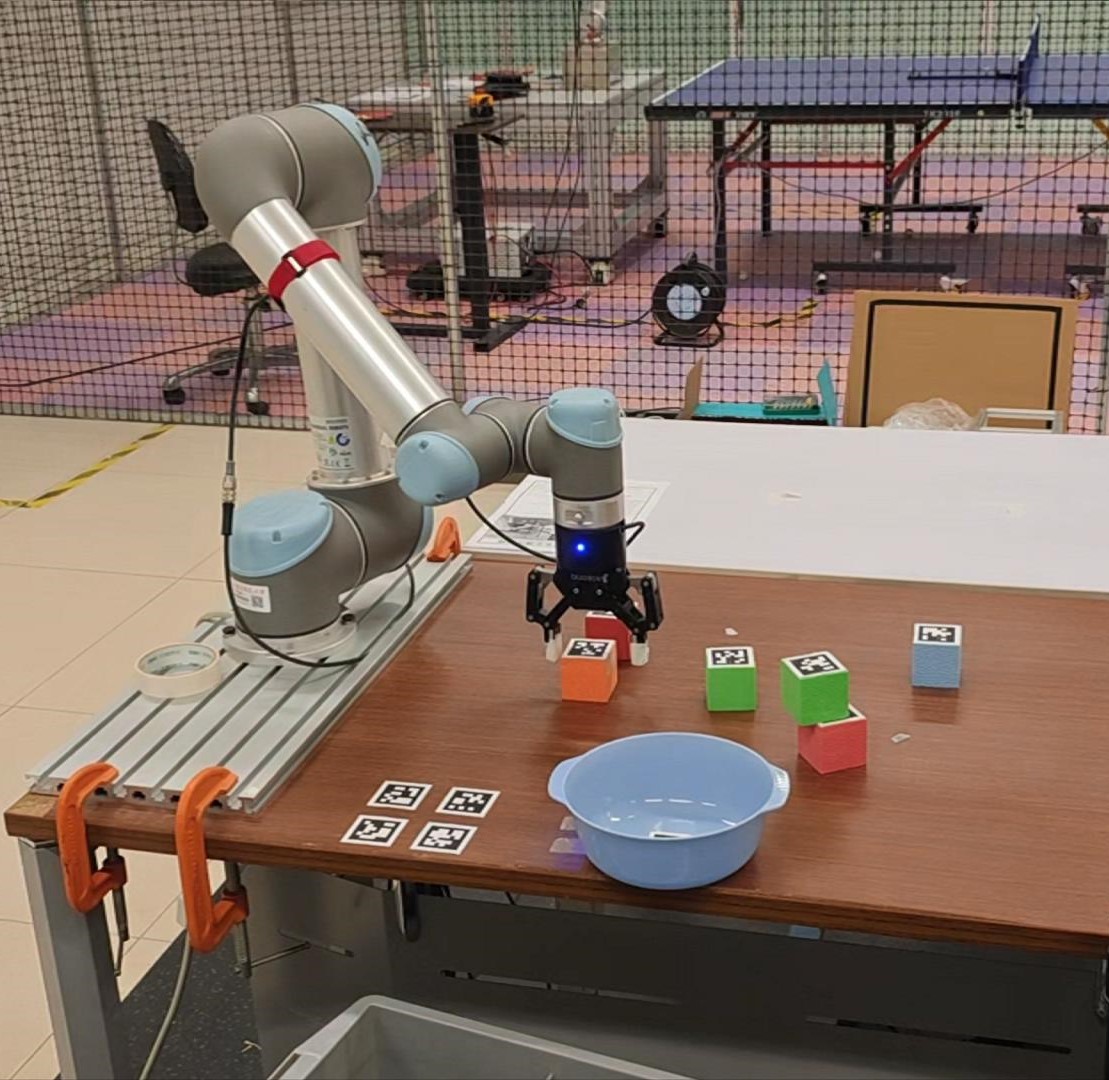}
  \caption{} 
  \label{fig:6cube4}
\end{subfigure}
\hfill
\begin{subfigure}{0.19\textwidth}
  \centering
  \includegraphics[width=\linewidth]{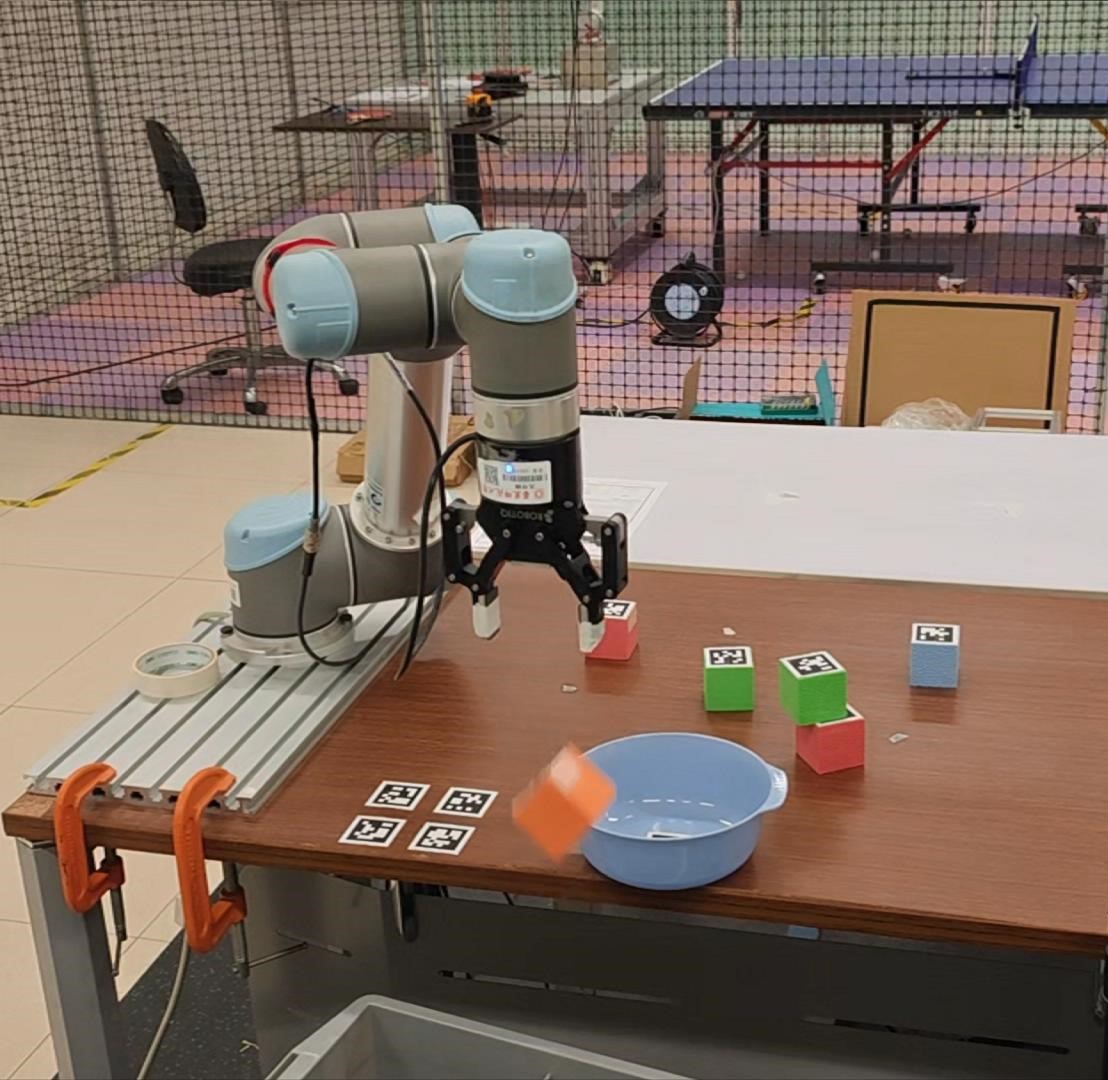}
  \caption{} 
  \label{fig:6cube5}
\end{subfigure}

\begin{subfigure}{0.19\textwidth}
  \centering
  \includegraphics[width=\linewidth]{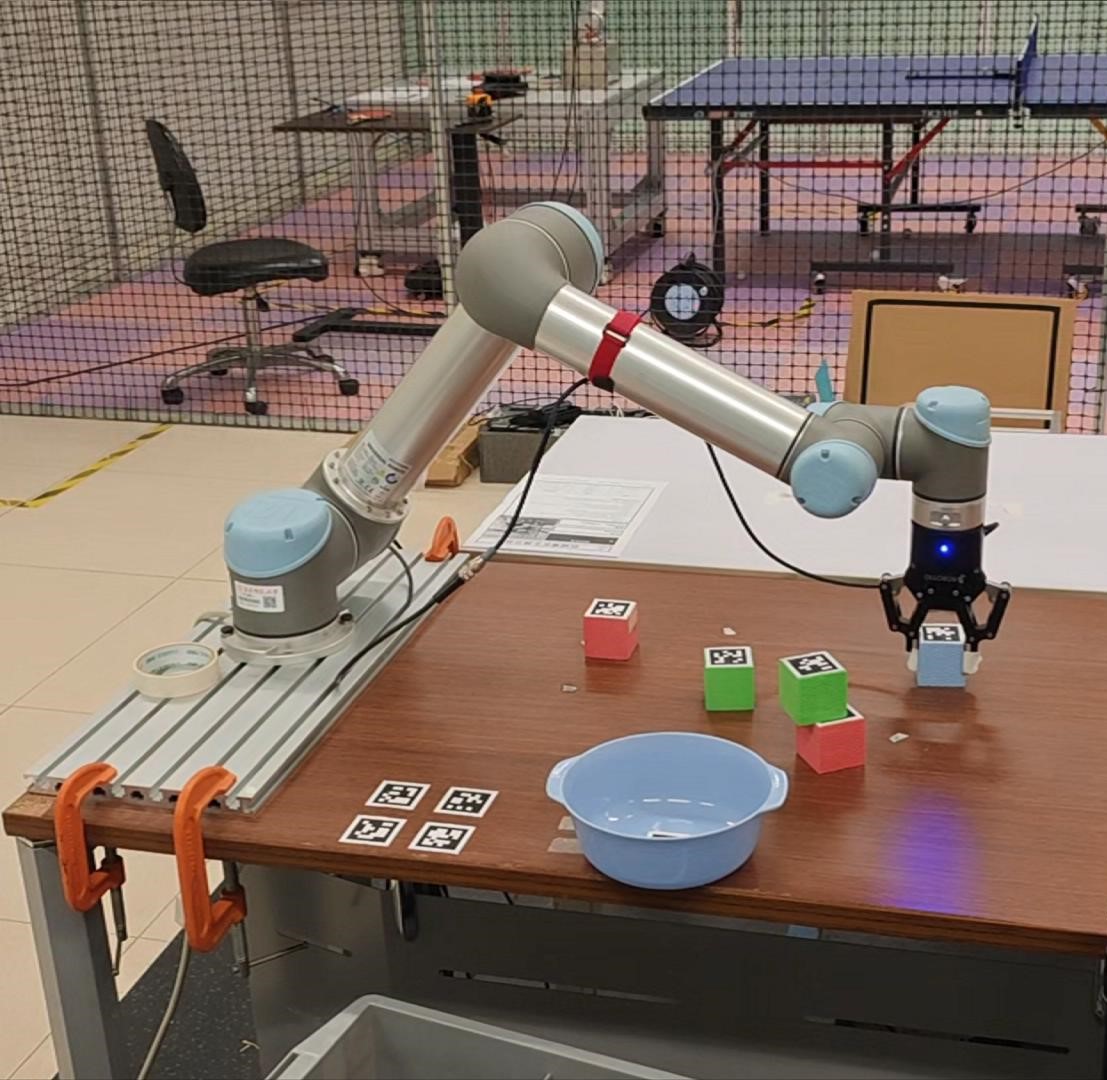}
  \caption{} 
  \label{fig:6cube6}
\end{subfigure}
\hfill
\begin{subfigure}{0.19\textwidth}
  \centering
  \includegraphics[width=\linewidth]{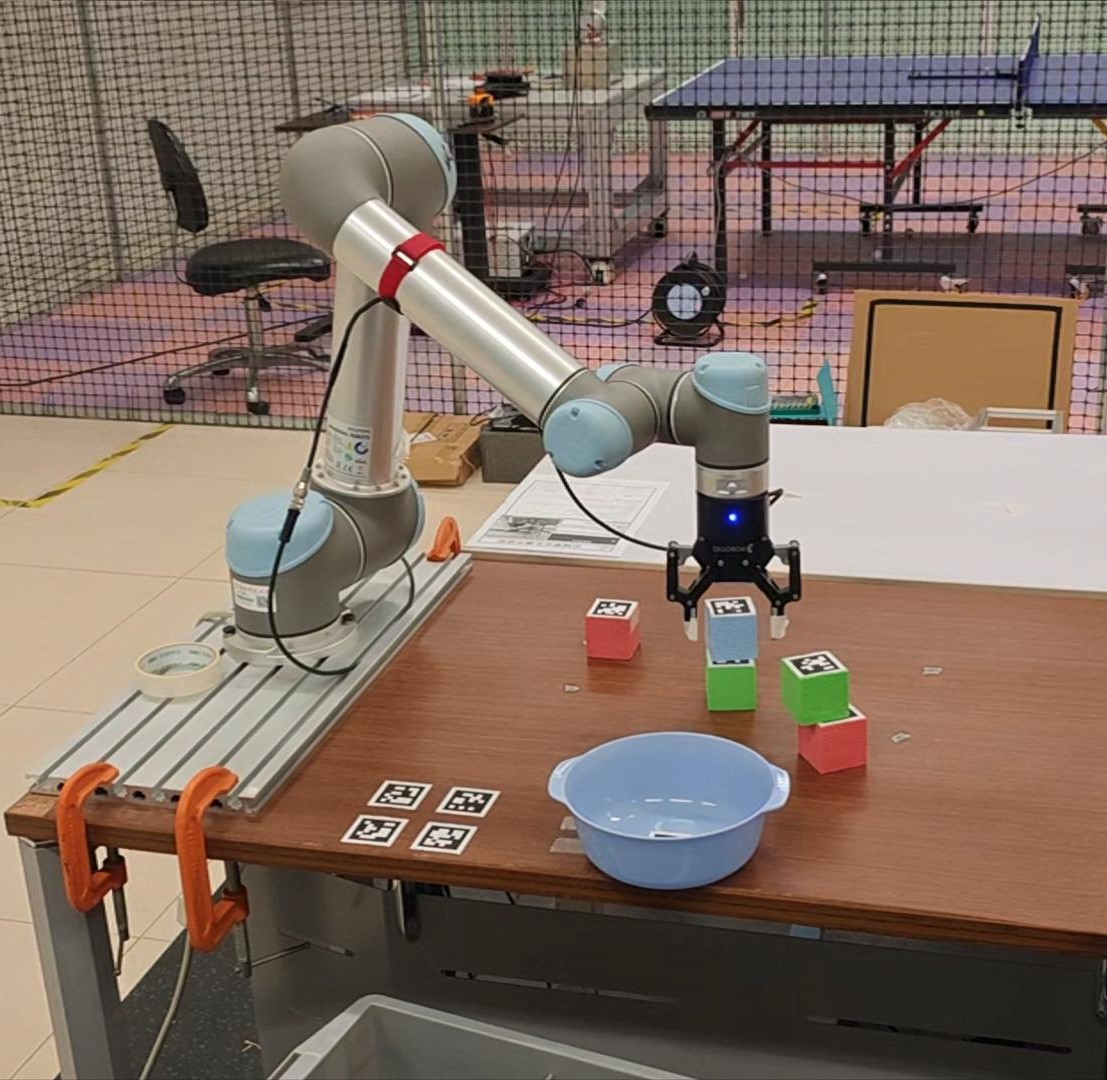}
  \caption{} 
  \label{fig:6cube7}
\end{subfigure}
\hfill
\begin{subfigure}{0.19\textwidth}
  \centering
  \includegraphics[width=\linewidth]{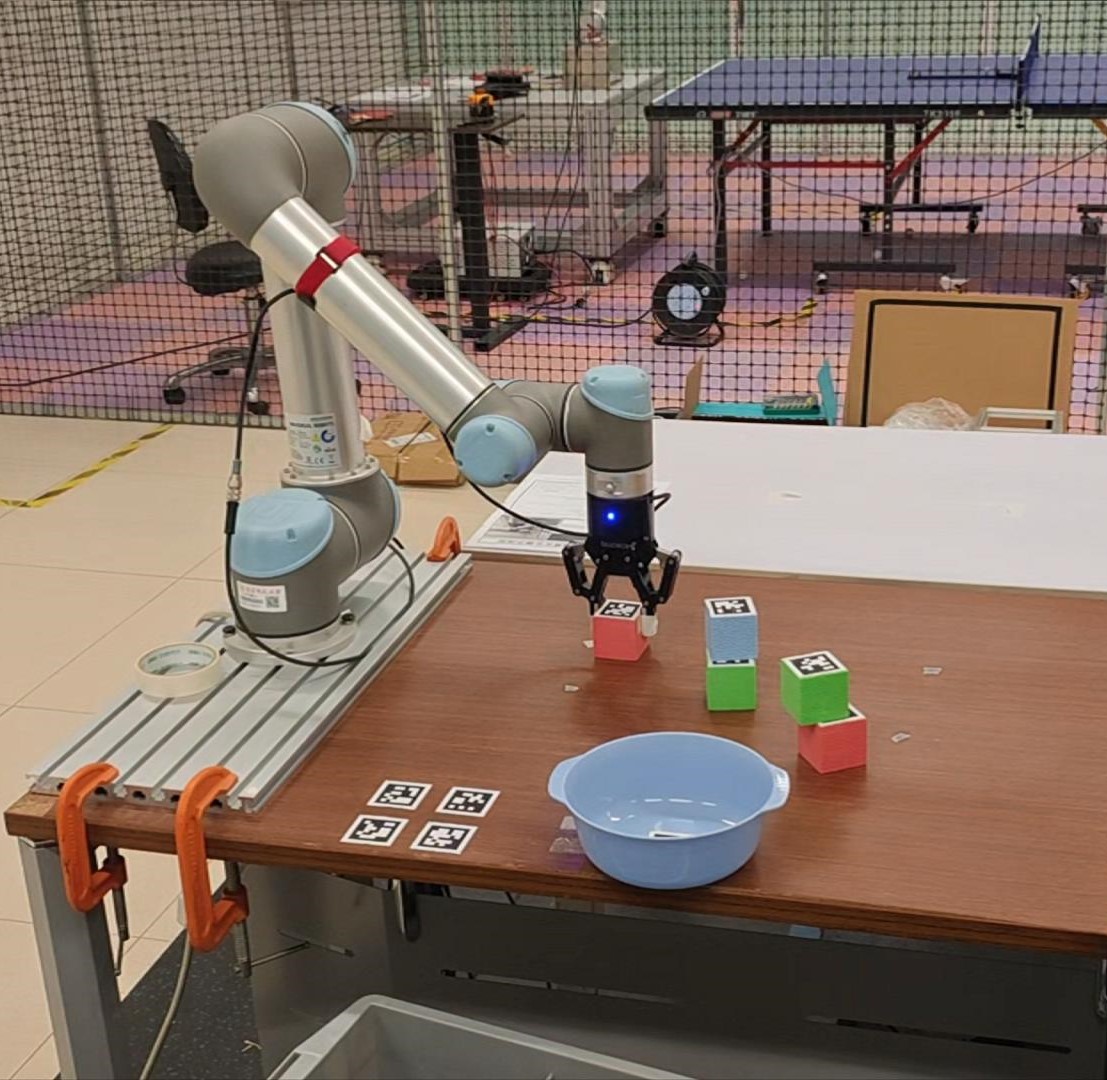}
  \caption{} 
  \label{fig:6cube8}
\end{subfigure}
\hfill
\begin{subfigure}{0.19\textwidth}
  \centering
  \includegraphics[width=\linewidth]{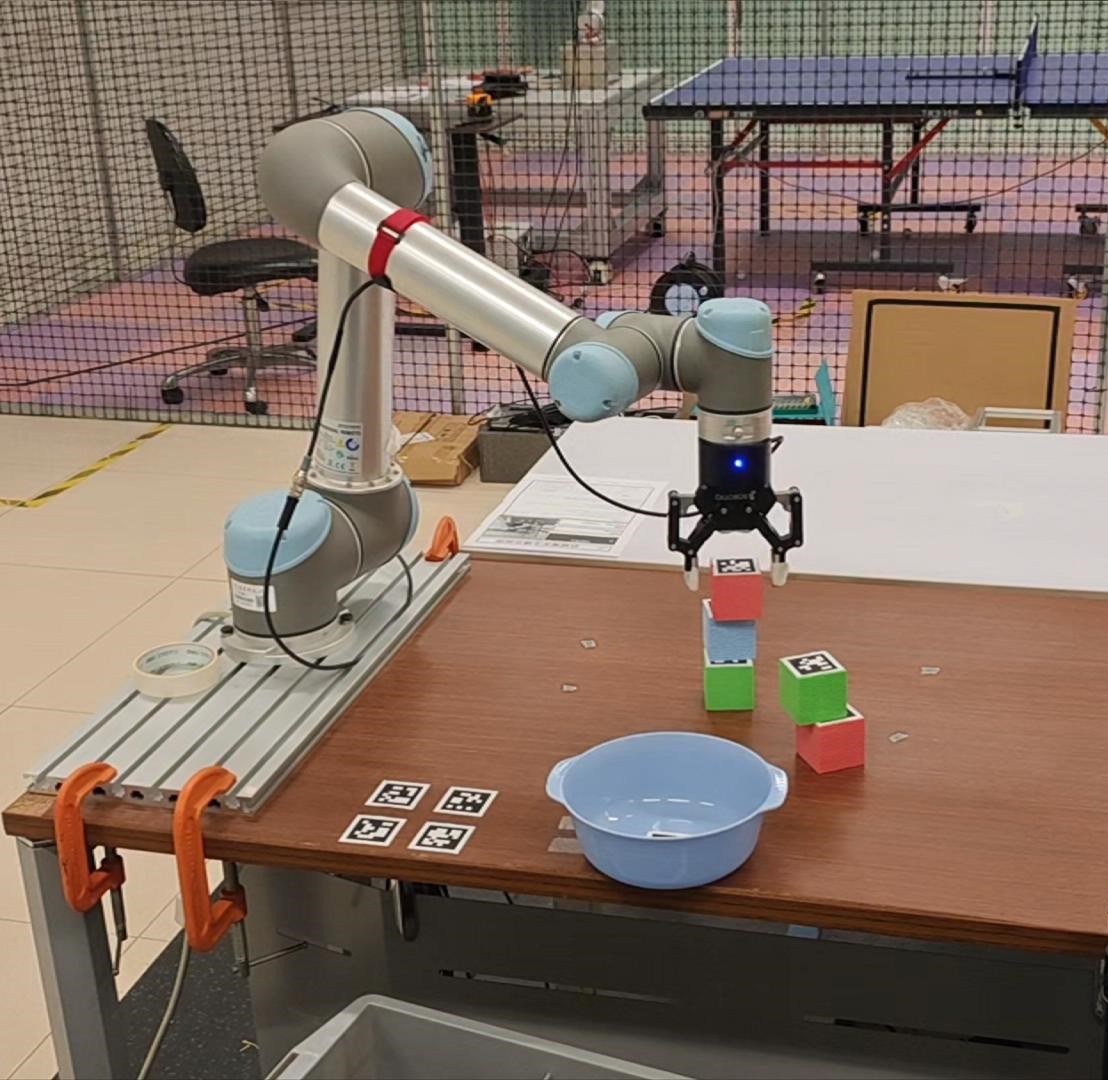}
  \caption{} 
  \label{fig:6cube9}
\end{subfigure}
\hfill
\begin{subfigure}{0.19\textwidth}
  \centering
  \includegraphics[width=\linewidth]{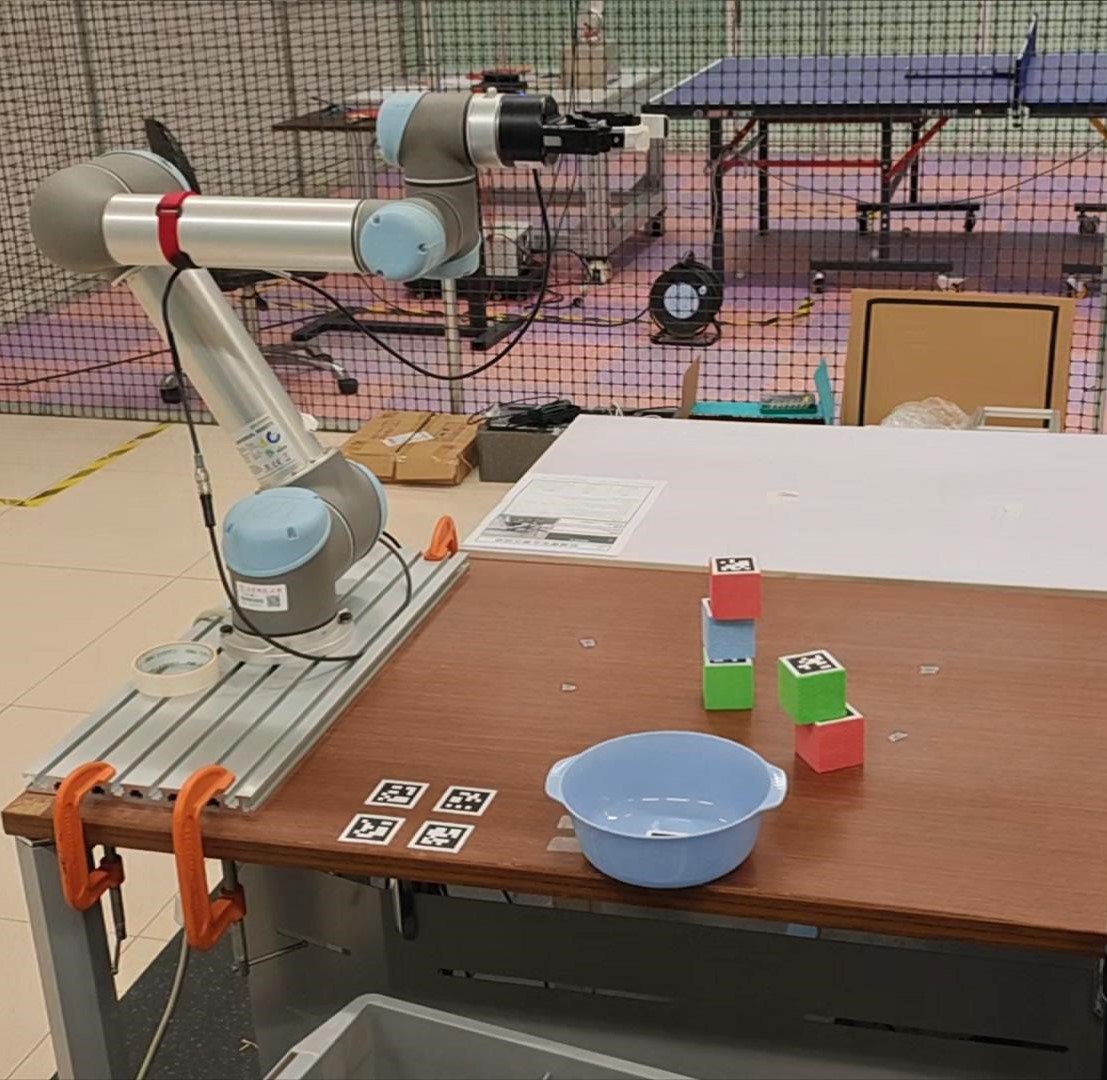}
  \caption{}
  \label{fig:6cube10}
\end{subfigure}

\caption{Sequence for the 6-cube task, depicted over ten stages: (a) starting, (b) picking up a green cube, (c) placing the green cube on top of a red cube, (d) picking up an orange cube, (e) dropping the orange cube into the sink, (f) picking up a blue cube, (g) placing the blue cube on top of another green cube, (h) picking up another orange cube, (i) placing the orange cube on top of the blue cube, (j) finishing.}

\label{fig:6cube}
\end{figure}

The ability to correct policy decisions based on human feedback is crucial to enhance the adaptability and robustness of autonomous systems. In this section, we present experiments assessing the effectiveness of human feedback in correcting policy errors generated by the LLM. Our objective is to demonstrate that, as task complexity increases, the performance of the LLM decreases, but that with the incorporation of human feedback, the success rate can be maintained.

We designed a tabletop rearrangement task involving several colored boxes placed on a table in front of the robot, with a bowl and a sink nearby. 
Each task prompt specifies instructions for rearranging the boxes, such as placing specific colors into the sink or bowl and stacking others. 
The difficulty of the task is determined by the number of boxes present in the scene, ranging from 2 to 8 boxes, with 5 unique tasks generated for each number of boxes. In total, 35 tasks were manually created.

For each trial, the LLM generates an action sequence and executes it on the real robot. If the action sequence is correct, the trial is considered a success; otherwise, human feedback is provided, the scene is reset, and the policy is executed again. 
We determine success by comparing the robot's actions with the known solution for each task. 
In cases where human feedback is provided, a second execution of the policy is performed, and we measure success in the same way.

\begin{figure}
    \centering
    \includegraphics[width=0.75\textwidth]{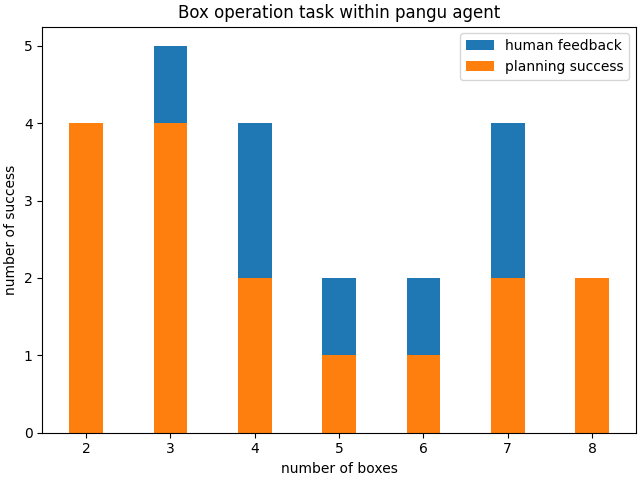}
    \caption{Results depicting policy correction through human feedback, where orange indicates task success with LLM planning and blue indicates task success after correction with human feedback. The graph illustrates a decrease in the success rate of the LLM as task complexity rises, countered by sustained success rates with the incorporation of human feedback for action sequence correction.}
    \label{fig:policy-correction-via-hf}
\end{figure}

Figure~\ref{fig:6cube} provides a sequence of images demonstrating our system executing a task, while
Figure~\ref{fig:policy-correction-via-hf} shows the results of our experiments. As expected, we observe a decrease in performance without human feedback as task difficulty increases. 
One example of human feedback is when the robot picks up the wrong cube, and the human indicates that it is not the correct cube and reiterates the correct task order.
Another example is when the robot fails to locate the target cube and raises an error due to the arm blocking the camera; the human provides the feedback as ``Please home the arm before looking for the cube.''
Conversely, the inclusion of human feedback generally leads to improved task success rates across varying levels of difficulty. 
However, we note that sometimes human feedback does not result in a successful task completion, highlighting potential limitations or challenges in the correction process.

Overall, our experiments demonstrate the effectiveness of human feedback in correcting policy errors generated by the LLM, thereby enhancing the adaptability and robustness of autonomous systems. 

\subsection{Updating the action library with imitation learning}

Enhancing robotic skill sets to include a variety of tasks is crucial, especially in household environments where the required actions may not be sufficiently covered by existing action libraries. To meet this need, we present an experiment that demonstrates the system's adaptability and capacity for continual learning. By showcasing its ability to learn new atomic actions from human demonstrations and seamlessly integrate them into the action library, we aim to highlight the system's versatility and its potential to evolve in response to user needs.

The experiment entails the demonstration of various tasks through kinaesthetic teaching, wherein a human guides the robot through the execution of specific actions. These actions include stirring, pouring, tossing the pan, wiping the table, seasoning food, and grating cheese.
Each demonstration is captured and represented using a DMP representation, facilitating the learning process. 
Subsequently, the human provides a descriptive label for each atomic action, enriching the action library with contextual information.

Following the addition of new atomic actions to the library, we challenge the system with a long-horizon task: ``make me pasta.'' 
Leveraging the updated action library, the system orchestrates a sequence of atomic actions to fully fulfill the task requirements.

This experiment highlights the system's proficiency in continual learning and adaptation, demonstrated through its ability to learn diverse atomic actions from human demonstrations and seamlessly incorporate them into the action library. The successful execution of the ``make me pasta'' task exemplifies the system's capability to effectively utilize learned atomic actions for complex, long-horizon tasks. Additionally, the inclusion of descriptive labels improves the action library's interpretability and usability, facilitating intuitive task specification by non-experts. This capability enhances the system's versatility and applicability in various real-world scenarios, emphasizing its potential as a valuable tool for intuitive robot programming and enhancing human-robot collaboration in household environments.

\begin{figure}[ht]
\centering

\begin{subfigure}{0.19\textwidth}
  \centering
  \includegraphics[width=\linewidth]{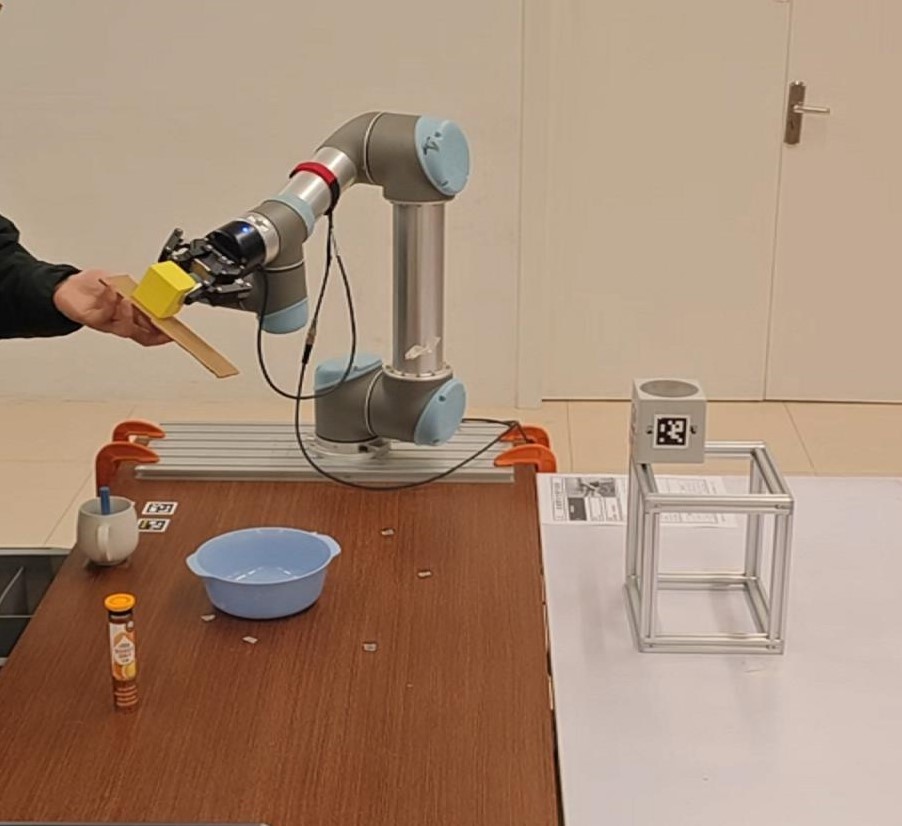}
  \caption{} 
  \label{fig:pasta1}
\end{subfigure}%
\hfill
\begin{subfigure}{0.19\textwidth}
  \centering
  \includegraphics[width=\linewidth]{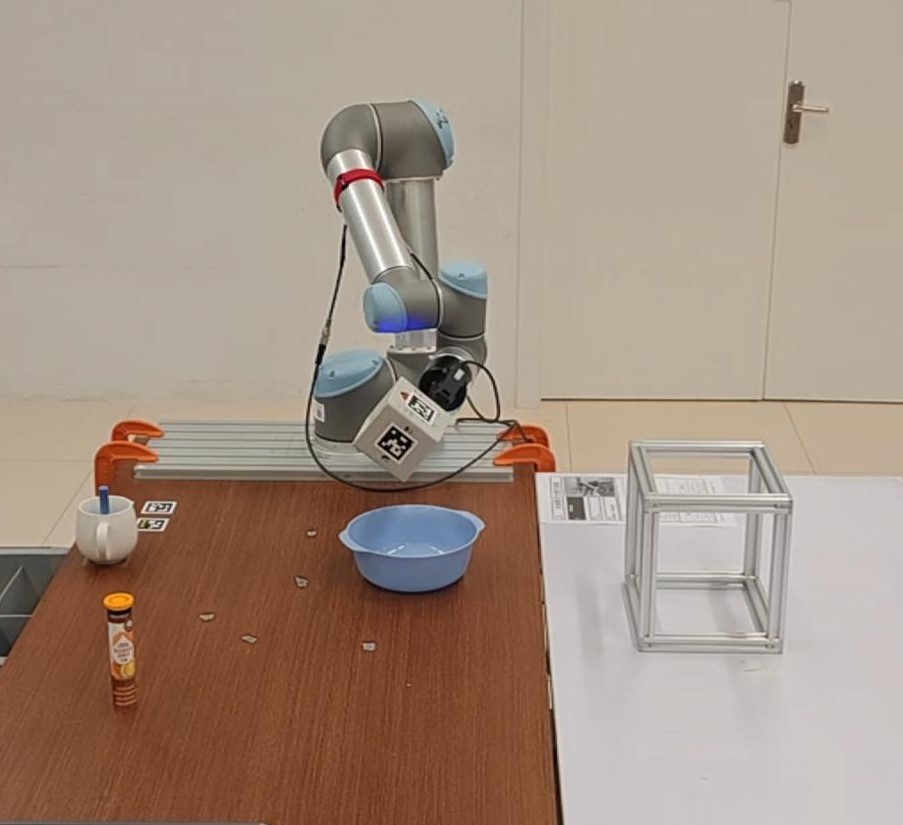}
  \caption{} 
  \label{fig:pasta2}
\end{subfigure}%
\hfill
\begin{subfigure}{0.19\textwidth}
  \centering
  \includegraphics[width=\linewidth]{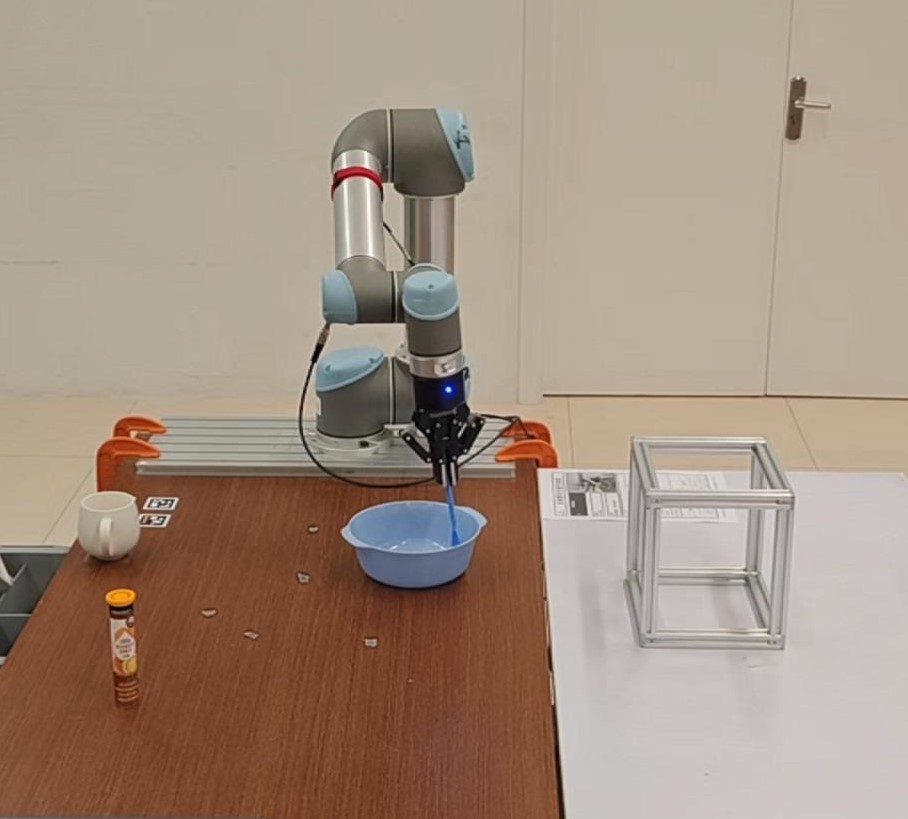}
  \caption{} 
  \label{fig:pasta3}
\end{subfigure}
\hfill
\begin{subfigure}{0.19\textwidth}
  \centering
  \includegraphics[width=\linewidth]{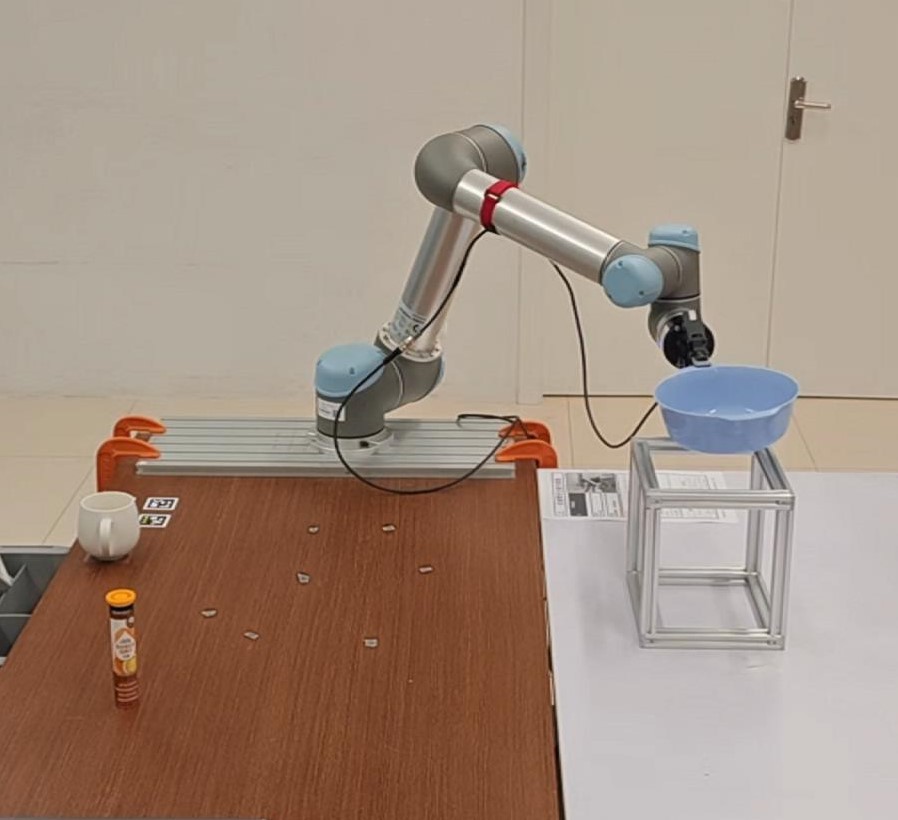}
  \caption{} 
  \label{fig:pasta4}
\end{subfigure}%
\hfill
\begin{subfigure}{0.19\textwidth}
  \centering
  \includegraphics[width=\linewidth]{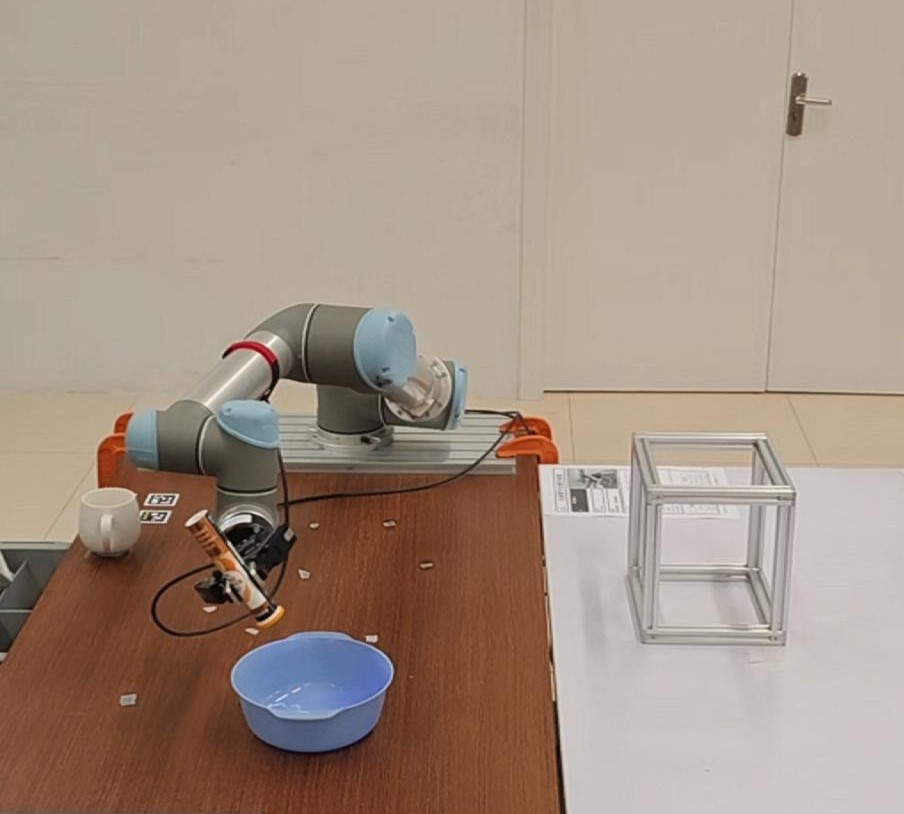}
  \caption{} 
  \label{fig:pasta5}
\end{subfigure}

\caption{Steps in the pasta-making process, depicted in five stages: (a) grating cheese, (b) pouring sauce, (c) stirring ingredients, (d) tossing the contents in the wok, (e) adding seasoning.}
\label{fig:pasta}
\end{figure}

\subsection{Adapting to a changing environment and continual learning}

The adaptability of robotic systems to dynamic and unpredictable changes in the environment is crucial for their effective operation in real-world scenarios. In this experiment, we aim to demonstrate the capability of our framework to adapt and recover from such changes through continual learning facilitated by human feedback. The primary goal is to showcase the system's ability to utilize human feedback to recover from unforeseen disruptions in task execution and subsequently learn from these experiences to handle similar changes in the future autonomously.

The experimental protocol is as follows.
\begin{enumerate}
    \item \textbf{Task Specification}: The system is tasked with ``pick and place the box,'' a common robotic manipulation task.
    \item \textbf{Action Sequence Generation}: Using the provided task description, the language model generates an action sequence for execution.
    \item \textbf{Execution and Disruption}: As the robot executes the task, the environment is intentionally perturbed by moving the target box, leading to a failure to grasp the object.
    \item \textbf{Human Feedback}: Upon observing the failure, the human provides corrective feedback, advising the system to ensure the box's proximity before grasping and then to retry the task.
    \item \textbf{Recovery and Adaptation}: Leveraging the feedback, the system adapts its approach and successfully completes the task, demonstrating its ability to recover from environmental changes through human-guided learning.
\end{enumerate}

To evaluate the system's capability for continual learning, a second trial is conducted under identical conditions, with the box moved simultaneously as in the previous trial. However, this time, the human feedback from the initial trial is incorporated into the task prompt. By doing so, we aim to assess whether the system autonomously applies the learned corrective action to handle similar environmental changes without human intervention.

The results of the experiment demonstrate the efficacy of our framework in adapting to changing environments and leveraging human feedback for continual learning. By successfully recovering from unforeseen disruptions and autonomously applying learned strategies in subsequent trials, the system showcases its resilience and ability to evolve through experience. These findings highlight the potential of our framework to enhance robustness and adaptability in real-world robotic applications, paving the way for more reliable and versatile autonomous systems.

\subsection{Remote supervisory control}

\begin{figure}[ht]
\centering

\begin{subfigure}{0.24\textwidth}
  \centering
  \includegraphics[width=\linewidth]{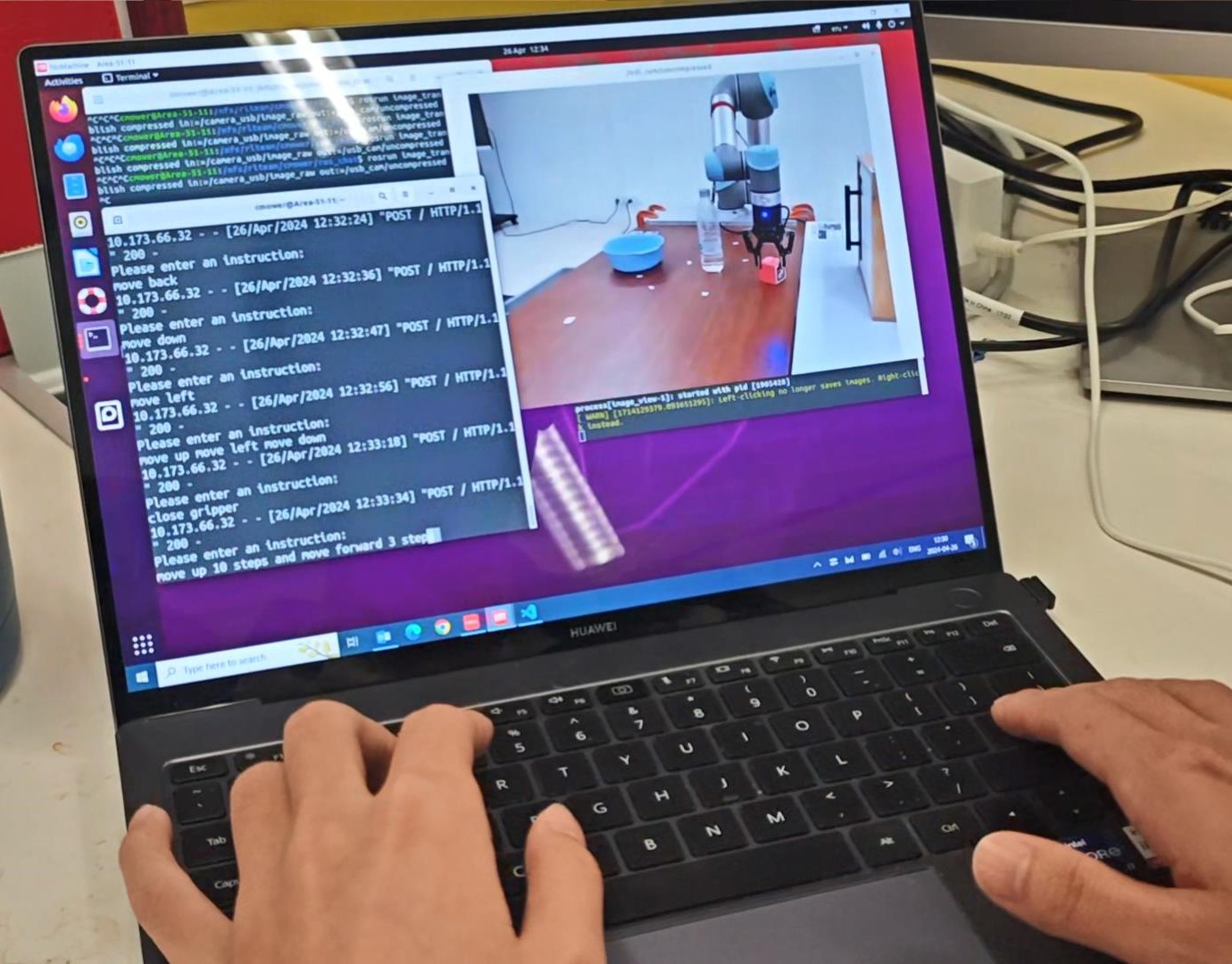}
  \caption{} 
  \label{fig:teleopllm4}
\end{subfigure}
\hfill
\begin{subfigure}{0.24\textwidth}
  \centering
  \includegraphics[width=\linewidth]{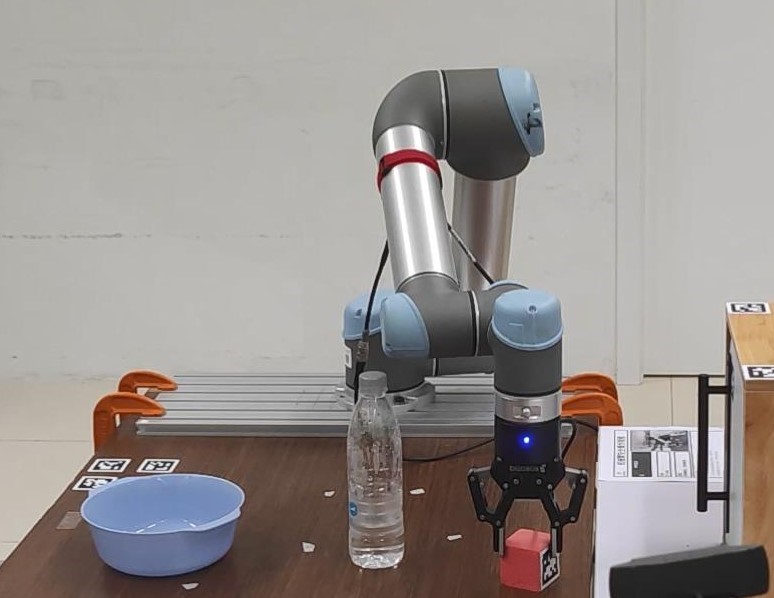}
  \caption{} 
  \label{fig:teleopllm1}
\end{subfigure}%
\hfill
\begin{subfigure}{0.24\textwidth}
  \centering
  \includegraphics[width=\linewidth]{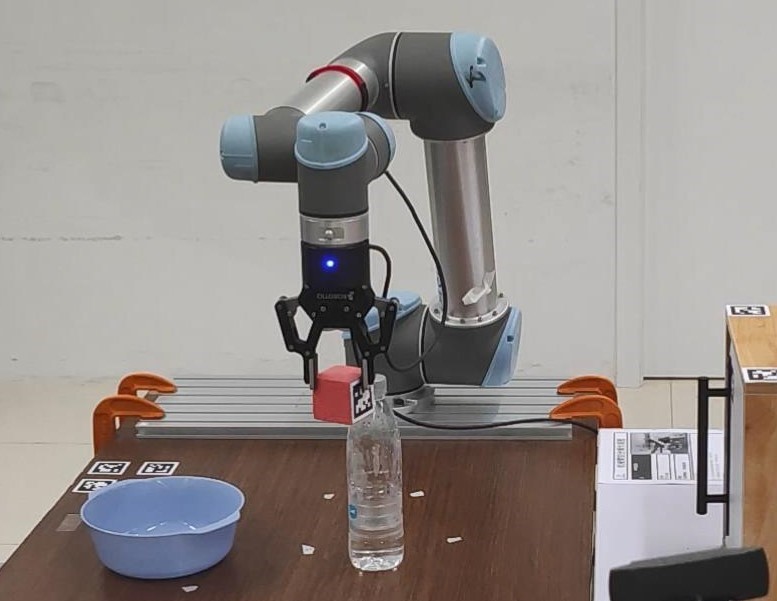}
  \caption{} 
  \label{fig:teleopllm2}
\end{subfigure}%
\hfill
\begin{subfigure}{0.24\textwidth}
  \centering
  \includegraphics[width=\linewidth]{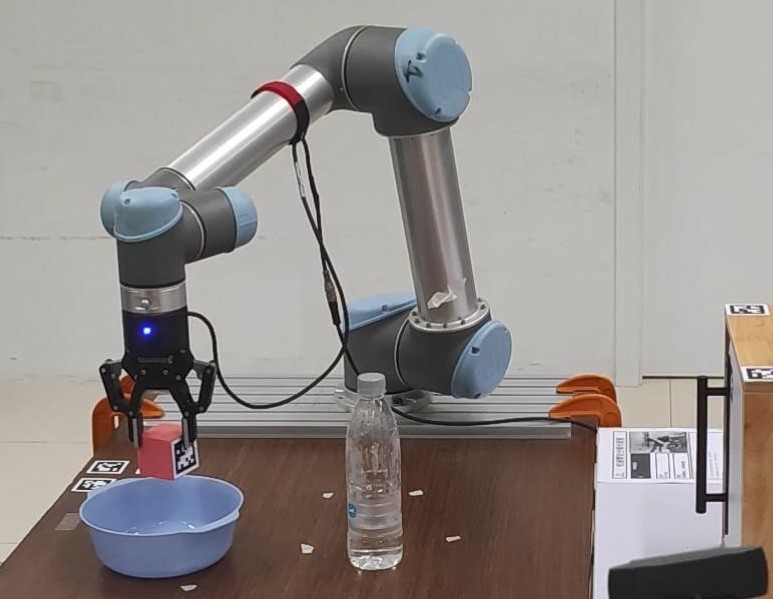}
  \caption{} 
  \label{fig:teleopllm3}
\end{subfigure}%
\hfill

\caption{Remote supervisory control using (a) language interfaces, depicted through continuous actions with two interfaces: (b) picking up a cube, (c) avoiding obstacles, (d) placing the cube in a bowl.}

\label{fig:teleop}
\end{figure}

The capability for remote supervisory control of robotic systems holds significant promise for a variety of applications, particularly in environments where direct human presence is impractical or unsafe, e.g.~\cite{Mower-RSS-21}.
Tasks such as offshore inspection and maintenance, search and rescue operations, mining activities, and space robotics often necessitate remote operations due to safety concerns and logistical challenges. 
Moreover, the integration of 5G technology further enhances the feasibility and effectiveness of remote control systems, offering high-speed, low-latency communication capabilities that are essential for near real-time interaction between operators and robots over vast distances.
For instance, handling obstacles in the environment, such as debris in search and rescue, may pose challenges for the robot's perception algorithms, requiring human intervention to guide the robot effectively.

In our experiment, conducted with the operator located in Europe and the robot system situated in Asia, participants were tasked with controlling the robot to perform pick-and-place manoeuvres with obstacles present in the scene. 
The action library provided to the operator included basic movement commands such as 
\texttt{move\_left}, 
\texttt{move\_right}, 
\texttt{move\_up}, 
\texttt{move\_down}, 
\texttt{move\_forward}, 
\texttt{move\_backward}, 
\texttt{open\_gripper}, and
\texttt{close\_gripper}.
The operator communicated commands to the robot using natural language, describing where they wanted the robot to move and manipulate objects in the environment.
A camera feed provided visual feedback to the operator, enabling them to perceive the scene remotely.
Remote supervisory control experiments using the language interface are shown in Figure~\ref{fig:teleopllm4}.

A significant lag in the system, measured at approximately 2–3 seconds, was observed due to the long distance between the operator and the robot system.
Despite this latency, participants successfully completed the assigned tasks using the language-based interface, demonstrating the effectiveness of our framework for remote supervisory control. 
\begin{figure}[t]
    \centering
    \includegraphics[width=\textwidth]{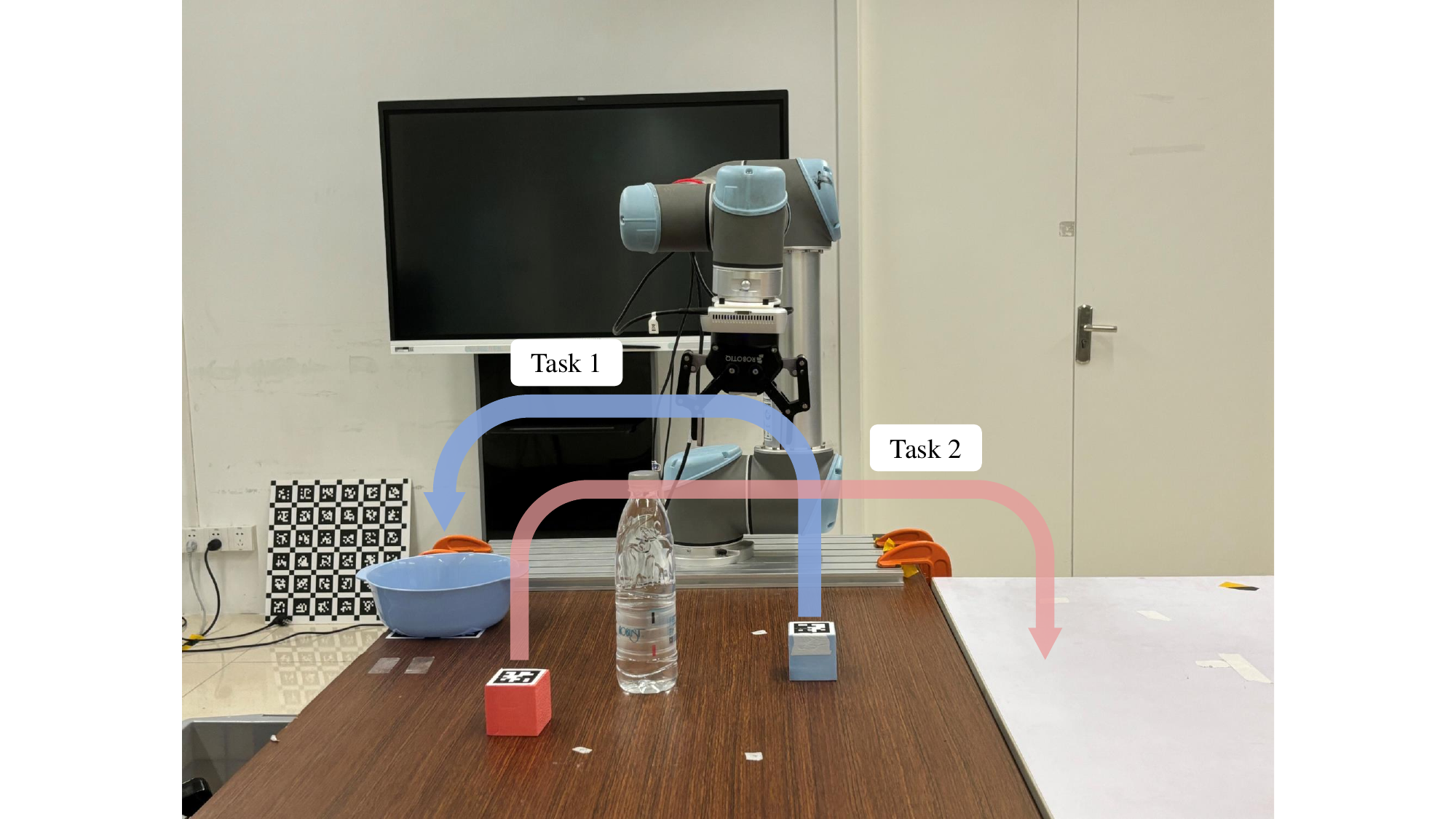}
    \caption{Experiment setup for the human study.}
    \label{fig:human_study}
\end{figure}

\begin{figure}[t]
    \centering
        \includegraphics[width=0.6\textwidth]{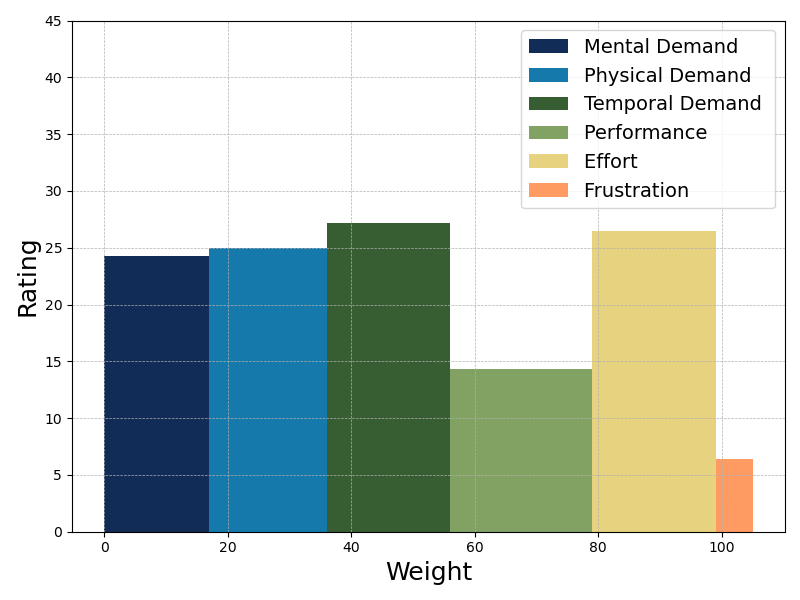}
        \label{fig:llm_tlx}
    \caption{Weighted NASA TLX results for remote supervisory control.}
    \label{fig:combined_tlx}
\end{figure}

Seven volunteers participated in the experiment. Each participant was required to complete two tasks. The first task involved picking up the blue box and placing it in the blue bowl, and the second task involved picking up a red box and placing it on the white area of the table while avoiding an obstacle in the middle, as shown in Figure \ref{fig:teleop} and \ref{fig:human_study}.
To evaluate the performance and usability of the human-robot interaction \cite{wan2023performance}, we recorded the time taken to complete each task. Furthermore, usability was assessed using the NASA Task Load Index (NASA-TLX) \cite{hart1986nasa} and pairwise subscale comparisons, which provided a comprehensive view of user experience across various dimensions.
Seven participants' results yielded valuable insights into the method's performance and usability, underscoring our framework as a practical tool for developing supervisory control using natural language in real-world scenarios. 
Detailed results are shown in Table \ref{table:task_time} and Figure \ref{fig:combined_tlx} for further analysis and discussion.

\begin{table}[]
\caption{Task time for the remote supervisory control experiment.}
\centering
\begin{tabular}{ccc}
\hline
\multicolumn{1}{l}{} & \multicolumn{2}{c}{\textbf{LLM}} \\ \hline
\multicolumn{1}{l}{\textbf{No.}} & \textbf{Task 1} & \textbf{Task 2} \\ \hline
1 & 5:09 & 9:40 \\
2 & 6:10 & 9:55 \\
3 & 5:06 & 7:31 \\
4 & 11:40 & 15:42 \\
5 & 5:05 & 4:51 \\
6 & 5:32 & 7:45 \\
7 & 4:27 & 7:37 \\ \hline
Avg & 6:09 & 9:00 \\ \hline
\end{tabular}
\label{table:task_time}
\end{table}

\section{Discussion}
The results of our experiments highlight the potential of this approach to simplify the interaction between non-expert users and robotic systems through natural language.
This section discusses the implications of our findings, the limitations and strengths of the current framework, and future directions for research.

\subsection{Performance observations in long-horizon tasks}
Alongside the promising outcomes, our extensive experiments have uncovered a range of interesting phenomena, showcasing both notable strengths and areas in need of improvement within the current framework.
These findings highlight the robust capabilities of our approach while also pinpointing challenges that guide further development.
This balanced perspective not only underscores the potential of our work but also clearly delineates paths for enhancing performance and reliability.

\subsubsection{Sensitive prompt}
During our experiments, we observed that minor variations in prompt wording significantly impact the model's output, revealing both vulnerabilities and strengths in its language processing capabilities.
For example, the model generated the correct action sequence when the prompt used ``another cube'' to indicate the second cube but struggled with ``the other cube.''
Similarly, discrepancies arose with the prompts ``put box A on box B'' versus ``put box A on top of box B,'' where the addition of ``on top of'' altered the generated actions.

Conversely, the model demonstrated robustness to other variations, such as replacing ``put box'' with ``move box'' and changing phrases from ``in the bowl'' to ``to the bowl.''
These instances did not affect the model's performance, indicating a certain level of linguistic flexibility.

Possible explanations for these observed behaviors might include:
\begin{itemize}
    \item \textbf{Training Data Variance:} The model's differential response could be attributed to the frequency and context of phrase occurrences in its training data. Phrases like ``move box'' and ``put box'' might appear interchangeably across diverse contexts, helping the model learn their equivalence.
    \item \textbf{Semantic Parsing Differences:} For the phrases that cause discrepancies, it may be that the model perceives ``on top of'' as indicating a more precise spatial relationship, thus requiring more specific handling in action generation.
    This points to a nuanced understanding of prepositions and their implications in task execution.
    \item \textbf{Context Sensitivity:} The model might be particularly sensitive to certain keywords or phrases that imply a change in action complexity or specificity, such as spatial relations that require precise positioning.
\end{itemize}

These findings underscore the need for enhanced natural language understanding within robotic AI systems. Improving the model's ability to generalize across varied linguistic inputs is crucial without losing contextual accuracy.
Enhancements should focus on better handling of semantic nuances and increasing the robustness against minor linguistic variations to develop more reliable and capable robotic assistants.

\subsubsection{Confusing example}
During our experiments, we encountered specific issues where including example action sequences in the prompts led to confusion for the model.
Notably, the model occasionally attempted to execute actions on objects mentioned in the example sequences, but that were not actually present in the real-world environment state.
This confusion stems from the model's inability to distinguish between illustrative examples provided for context with current environment observation and actual commands.

This issue highlights a significant challenge in the design of natural language interfaces for robotic systems: the need for improved parsing and contextual understanding.
The model's current parsing mechanisms may not effectively differentiate between descriptive content (intended to enhance understanding or provide background information) and imperative content (direct commands). 

Possible strategies to mitigate this issue could include:
\begin{itemize}
    \item \textbf{Contextual Tagging:} Implementing a tagging system in the training phase where example actions and real commands are tagged differently could help the model learn to distinguish between these types of content more effectively.
    \item \textbf{Enhanced Semantic Analysis:} Developing more advanced semantic analysis capabilities that can interpret the context and intent behind each phrase more accurately.
    This could involve deeper training on linguistic cues that indicate hypothetical or illustrative scenarios versus actionable instructions.
    \item \textbf{User Feedback Integration:} Incorporating a feedback loop where the system asks for confirmation or clarification when it detects potential ambiguities in the prompt could prevent incorrect actions based on misinterpretations.
\end{itemize}

Addressing these parsing and contextual understanding challenges is crucial for advancing the usability and reliability of natural language interfaces in robotic systems.
By enhancing the model’s ability to discern and segregate different types of linguistic inputs accurately, we can significantly reduce errors and improve the system's overall performance in real-world tasks.

\subsubsection{Robust long-term planning}
An interesting aspect of our experiments was evaluating the model's long-horizon planning capabilities. Typically, the success rate of tasks is expected to decrease with an increase in the number of steps involved due to the accumulation of potential errors at each stage.
However, our findings challenge this expectation.
The model demonstrated a notable consistency in success rates across experiments involving 4 to 8 cubes.
This observation suggests that the model possesses robust error-handling and planning capabilities that maintain performance even as task complexity increases, as shown in Figure \ref{fig:policy-correction-via-hf}. 

These results are encouraging for the deployment of this framework in scenarios requiring complex sequential task execution, highlighting its potential reliability and effectiveness in practical applications.

\subsection{Enhancing policy correction via human feedback}
Targeted human feedback has shown the potential to mitigate this degradation by correcting erroneous policy decisions dynamically. We noticed that due to the variety of mistakes made by the LLM, there is no one-size-fits-all feedback correction. 

\subsubsection{Feedback implementation}
Our findings underscore that not all feedback is equally effective. Specific, actionable feedback that directly corrects the decision-making process or clarifies the task's objectives tends to result in more favourable outcomes. For instance, if the LLM incorrectly sequences the actions (e.g. attempting to place a cube before picking it up), repeating the command with the correct sequence often resolves the error, effectively reprogramming the LLM's task strategy.

Also, the human feedback needs to indicate the correction action or action sequence instead of only pointing out the mistake itself. 
For example, if the LLM erroneously attempts to place a blue cube on a green cube before securing the blue cube, the provided feedback would explicitly instruct to ``Pick up the blue cube first, then place it on the green cube.'' instead of ''You picked up the cube in a wrong order.''

\subsection{Usability for remote supervisory control}

From the remote supervisory control experiment, we observed notable phenomena related to the human-robot interface. Overall, we noticed that participants experienced similar mental, physical, and temporal demands during the operation, as shown in Figure \ref{fig:combined_tlx}. This suggests that the system maintains a balanced workload distribution.

\subsubsection{Stability of LLM}

During the experiment, we found that the LLM interface exhibited stability issues. Operators' freedom to input any command increased the likelihood of ambiguous instructions and typographical errors. 
For instance, operators have placed commands outside the giving action, including ``go but avoid the water'' and ``move the hand as close to the table as possible''. This highlights the need for further research to improve the robustness of LLMs in managing ambiguous inputs during robotic operations.

Also, even when commands were entered correctly, there was still a small chance of execution errors.
There were instances where the robot executed a command twice, repeated a command continuously without stopping, or misinterpreted the number of steps required by a command.

One participant left the following feedback:`` Language is easier to use due to the ability to execute sequences, although occasionally the sequence was not executed entirely correctly.''





\subsubsection{Limitations of the experiment}

In addition to the LLM being tested, visual feedback significantly impacts the system's performance and workload.
Participant feedback indicates that video resolution plays a more crucial role than time delay.
Furthermore, only providing a single camera view for the operator in the experiment made it challenging to accurately determine the distance to an object.
However, with higher resolution, participants were able to infer the relative position of other objects through shadows and reflections.

Additionally, mistakes can significantly impact task duration. The most common error is dropping the cube outside the bowl, requiring the robot to pick up the cube again, which is time-consuming.
In some cases, when the cube is dropped very close to the bowl, it requires fine movement to get the end-effector to the right position.

Another common mistake is perspective reversal. However, the participant usually can recover quickly from the mistake, and it only has a small impact on task time.

\section{Related work}

This section reviews various approaches that enhance the interaction between robots and human language, focusing on task planning through action composition with the integration of language models and robotic systems.
The comparative analysis presented in Table \ref{table:related_work} highlights distinct contributions in the existing approaches within the domain of robotic task execution using language models, including utilising open-source model, correcting mistakes on the fly with human and environment feedback, validating with real robot experiment, capability for ROS interface and fine-tuning, and integrating multimodal.

Kim et al. \cite{kim2024understanding} conducted a user study on LLM-powered human-robot interaction. Their findings indicate that LLM-powered robots excel in tasks involving connection-building and deliberation but face challenges in logical communication and inducing anxiety.

\begin{table}
\caption{Comparison of related works on the Use of Open-Sourced Models, Human Feedback, Real Robot Experiments, and ROS Integration.}
\label{table:related_work}
\centering
\begin{adjustbox}{center}
\begin{tabular}{@{}>{\raggedright}p{0.35\linewidth}>{\centering}p{0.07\linewidth}>{\centering}p{0.05\linewidth}>{\centering}p{0.07\linewidth}>{\centering}p{0.05\linewidth}>{\centering}p{0.07\linewidth}>{\centering}p{0.05\linewidth}>{\centering\arraybackslash}p{0.05\linewidth}>{\centering\arraybackslash}p{0.07\linewidth}@{}}
\toprule
& \rotatebox{90}{\textbf{Open-Source Model}} & \rotatebox{90}{\textbf{Human Feedback}} & \rotatebox{90}{\textbf{Environment Feedback}} & \rotatebox{90}{\textbf{Real Robot Experiment}} & \rotatebox{90}{\textbf{Remote Control}} & \rotatebox{90}{\textbf{ROS Capability}} & \rotatebox{90}{\textbf{Fine-Tuning Capability}} & \rotatebox{90}{\textbf{Multimodal Integration}} \\ 
\hline
\textbf{Ours} & \checkmark & \checkmark & \checkmark & \checkmark & \checkmark & \checkmark & \checkmark & WIP \\
Btgenbot \cite{izzo2024btgenbot} &   &   &   & \checkmark &   & \checkmark & \checkmark &   \\
GLM2FSA \cite{yang2023large} &   &   & \checkmark &   &   &   &   &   \\
Socratic models \cite{zeng2022socratic} &   &   &   &   &   &   &   & \checkmark  \\
Chain of code \cite{li2023chain} &   &   &    &   &   &   &   &   \\
To help or not to help \cite{tanneberg2024help} &   &   & \checkmark & \checkmark &   &   &   &   \\
Kwon et al. \cite{kwon2023language} &   &   &   & \checkmark &   &   &   &   \\
Statler \cite{yoneda2023statler} &   &   &   &  \checkmark &    &   &   &   \\
Embodiedgpt \cite{mu2023embodiedgpt} & \checkmark  &   &   &   &   &   & \checkmark & \checkmark \\
Huang, Abbeel, et al.  \cite{huang2022language} &  &   &   &   &   &   &   &   \\
Yell at your robot \cite{shi2024yell} &   & \checkmark & \checkmark & \checkmark  &  &   & \checkmark & \checkmark \\
Rt-h \cite{belkhale2024rth} &   &   &   & \checkmark &  &   &  \checkmark & \checkmark \\
Vima \cite{jiang2023vima} & \checkmark  &   &   &   &  &   & \checkmark  & \checkmark  \\
Natural language as policies \cite{mikami2024natural} &   &   &   &  &   &   &  & \checkmark \\
Clipswarm \cite{pueyo2024clipswarm} & \checkmark &   &   &  &   &   &   & \checkmark   \\
Progprompt \cite{singh2022progprompt} &   & \checkmark &   & \checkmark &   &   &   &   \\
Llm-brain \cite{lykov2023llmbrain} & \checkmark &   &   &   &   &   & \checkmark & \checkmark\\
Cao and Lee\cite{cao2023robot} &   &   &   &   &    &   &   &   \\
Grounded decoding \cite{huang2023grounded} &   &   &   &  \checkmark &  &   &    & \checkmark \\
Silver, Tom, et al. \cite{silver2023generalized} &   &   &   &   &   &   &   &   \\
CLMASP \cite{lin2024clmasp} &   &   &   &   &   &   &   &   \\
InterPreT \cite{han2024interpret} &   & \checkmark &   & \checkmark &   &   &   &   \\
LLM-Planner \cite{song2023llm} &   &   & \checkmark  &   &   &   &   &   \\
OLAF \cite{liu2023interactiverobotlearningverbal} &   & \checkmark  &   &  \checkmark &   &   & \checkmark  &   \\
LaMI \cite{wang2024lami} & & & & \checkmark & & \checkmark & & \checkmark \\
\bottomrule
\end{tabular}
\end{adjustbox}
\end{table}

\subsection{Action composition from language}
The integration of language models with robotic systems for task planning through action composition has been explored in various studies.
Zeng et al. \cite{zeng2022socratic} developed Socratic Models to enhance multimodal capabilities without the need for fine-tuning. Li et al. \cite{li2023chain} introduced Chain of Code, which leverages code execution to bolster reasoning processes. Kwon et al. \cite{kwon2023language} demonstrated the use of language models to predict robot end-effector poses based on visual inputs and task descriptions, showcasing direct application in robotic control. Silver et al. \cite{silver2023generalized} explored how closed-sourced models like GPT-4 can generate Python programs for task planning within Planning Domain Definition Language (PDDL) domains, a framework with a long history in the field of automated planning and scheduling.

Song et al. \cite{song2023llm} introduced the LLM-Planner for embodied agents, focusing on few-shot grounded planning to leverage LLMs, and emphasized the need for dynamic planning adaptability in complex environments.
This work aligns closely with traditional symbolic planning \cite{4141034} but introduces a novel integration of natural language processing, addressing both the symbolic and dynamic aspects of task planning. Wang et al. \cite{wang2024lami} developed LaMI, an LLM-based system designed to enhance multi-modal human-robot interaction. This system integrates high-level linguistic guidance, atomic actions, and multi-modal expressions to regulate robot behavior.

Izzo et al. \cite{izzo2024btgenbot} and Yang et al. \cite{yang2023large} take a structured approach by translating natural language into behavior trees and state machines, respectively, which guides robotic behavior in a fixed and somewhat rigid framework.
In contrast, our work emphasizes a more dynamic and flexible generation of executable action sequences directly from natural language prompts.
This not only simplifies the user interface but also enhances the adaptability of the system to a wider range of tasks and environments without the constraints of pre-defined structures of behavior trees or state machines.
Our method also corrects its mistakes based on human and environmental feedback, crucial for practical applications in dynamic environments, setting it apart from these more static and predefined approaches.

\subsection{Incorporating human feedback}
Most current systems operate in an open-loop configuration without the capability for error recovery. Shi et al. \cite{shi2024yell} introduce a framework where robots anticipate failures and proactively request human help to refine their task execution. Han et al. \cite{han2024interpret} develop a system that allows robots to explain their actions and receive corrective feedback from users, enhancing transparency and trust. Singh et al. \cite{singh2022progprompt} create an interactive prompting system that uses structured feedback to correct robot tasks, catering especially to technical and educational settings. Liu et al. \cite{liu2023interactiverobotlearningverbal} introduced OLAF (Operation-relabeled Learning with Language Feedback), a system that learns robot policies interactively using verbal corrections. It updates the robot’s visuomotor neural policy based on verbal feedback to avoid repeating mistakes. Our framework enables real-time interaction and task execution, facilitating natural user interaction between steps without requiring prior coding knowledge.

\subsection{Utilizing open-source models}
The use of open-source models in robotics research greatly facilitates accessibility and reproducibility, essential for advancing the field. Open-source frameworks allow researchers and developers to replicate studies and verify results, enhancing the credibility and scalability of the technologies developed. Mu et al. \cite{mu2023embodiedgpt} and Huang \cite{huang2023grounded} harness various open-source language models to drive their research, while Pueyo et al. \cite{pueyo2024clipswarm} utilize the multimodal capabilities of the CLIP model to explore new robotic functionalities.

Our approach builds on these foundations by exclusively employing open-source models, which ensures that our methodologies are transparent and easily accessible to the robotics community. This commitment supports more stable development environments and reproducibility of results. Moreover, by integrating these models with real-world feedback mechanisms, our framework enhances the dynamic control capabilities necessary for practical applications, addressing critical gaps observed in previous studies like those by Cao and Lee \cite{cao2023robot}, which lack empirical validation on actual robotic platforms. Thus, our work not only leverages the foundational benefits of open-source software to ensure reproducibility and enhance community engagement but also pushes the boundaries of what these models can achieve in real-world settings.

\subsection{Experiment on real robots}
Experimenting with real robots is a crucial step in robotics research, as it allows for the validation and refinement of theoretical models. Several studies have implemented their methodologies on real robots, but the integration with widely used platforms like the Robot Operating System (ROS) remains limited. Tanneberg et al. \cite{tanneberg2024help}, Kwon et al. \cite{kwon2023language}, and others have demonstrated practical applications, yet the potential for easier study and development through ROS integration is largely untapped. Our framework aims to fill this gap, providing a robust solution that combines the benefits of open-source models, human feedback, and easy deployment. Furthermore, contrasting with simulation-only studies such as Pueyo et al. \cite{pueyo2024clipswarm}, our framework emphasizes practical, real-world testing to ensure operational reliability and scalability.

\section{Use cases}

In this section, we report use-cases of our framework in the wild.

\subsection{Robotics Humanoid Kitchen Demonstration}


The proposed framework has been recently adapted to demonstrate their humanoid robot working in a kitchen setting.
The robot was tasked with cooking a meal based on user input, 
i.e. the human asked the robot to cook a particular dish.
The system hardware includes 
the humanoid robot, 
a cooking machine, and
several pre-prepared ingredients placed in bowls.
The humanoid robot is equipped with an Intel Realsense RGB-D camera on its head.

Atomic actions were developed for both the humanoid robot and the cooking machine system. 
The cooking machine actions are pre-set by the manufacturer, e.g. ``heat pan to \{TEMPERATURE\}'' where \{TEMPERATURE\} is the desired temperature.
In the case of the humanoid robot, atomic actions are developed solely with imitation learning.
For example, a control policy was learned for the atomic action with textual description ``put the \{INGREDIENT\} in the bowl'' where \{INGREDIENT\} is replaced with an ingredient such as ``broccoli''.
Several demonstrations of each atomic action are provided via a teleoperation interface. 
Once a database of demonstrations is collected, a policy is learned that maps the textual description and RGB image to the robot's end-effector displacement.
Subsequently, an inverse-kinematics controller translates these end-effector displacements into specific joint commands.
Once refined, this learned control policy is systematically integrated into the library as a predefined atomic action.

The voice input from the user is recorded using a microphone and converted into text which is used as the task description.
Given the task description, the language model generates an appropriate sequence for the atomic actions that is then executed on the system.
Future development plans are to increase the atomic library, improve capabilities of the humanoid system, and replace the language model with a vision-language model. 

\subsection{Robotics Air Hockey Challenge}

\begin{figure}[t]
    \centering
    \includegraphics[width=0.8\textwidth]{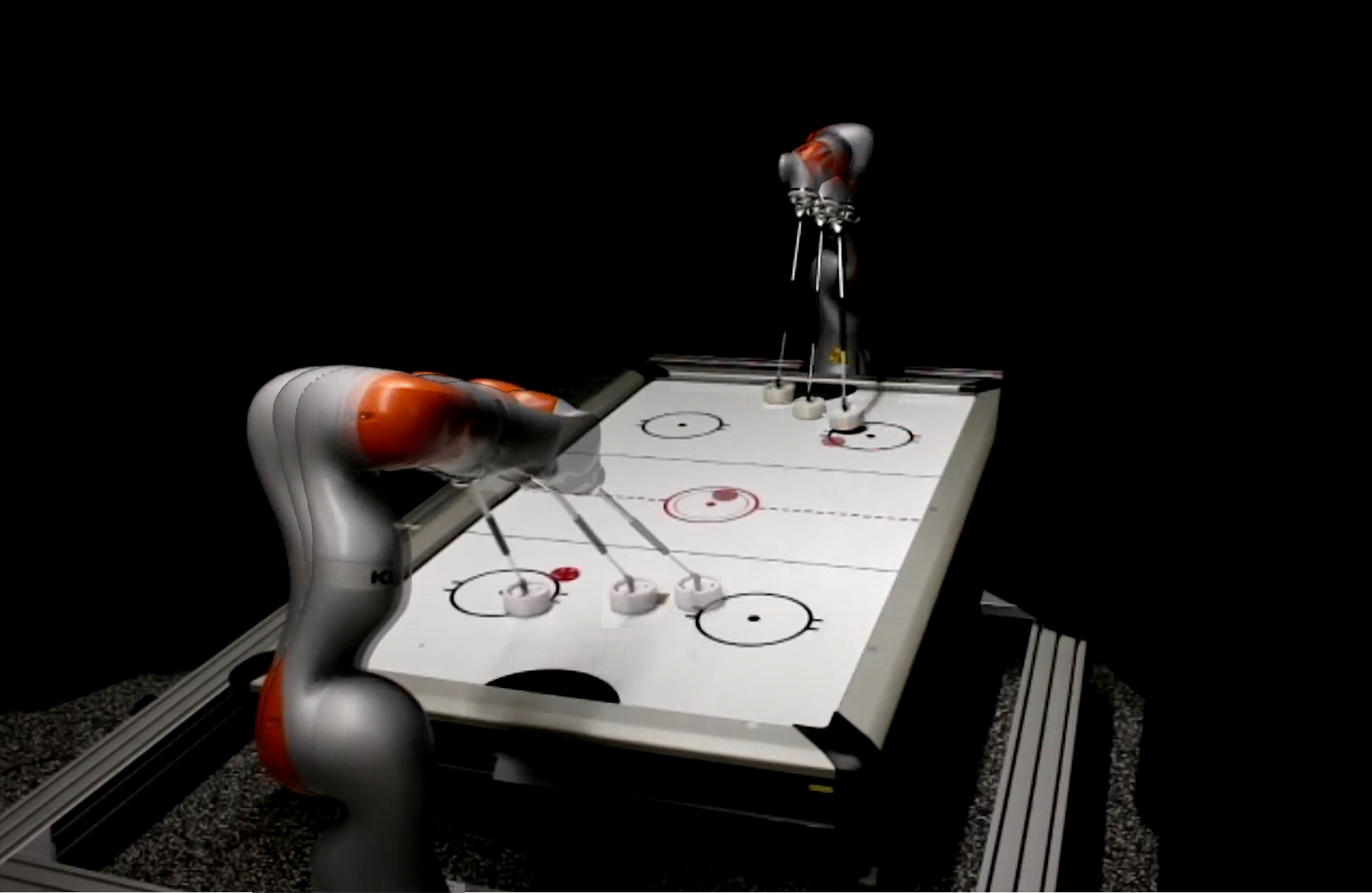}
    \caption{The setup for the Robot Air Hockey Challenge with two KUKA IIWA robot arms.}
    \label{fig:air-hockey}
\end{figure}

Bridging the simulation-reality gap is crucial for achieving high-performing embodied intelligence. 
Deploying learning approaches on real-world robots requires addressing practical challenges such as 
disturbances, 
observation noise, 
safety, 
model mismatches, 
delays, 
actuator limitations, 
physical feasibility, and 
limited real-world interactions. 
Additionally, future robots must 
dynamically react to their environments, 
execute agile movements, and 
engage in long-horizon planning.
The Robot Air Hockey Challenge, shown in Figure \ref{fig:air-hockey}, was developed as a collaborative platform for researchers to tackle a realistic robotic task. 
Teams designed and built air hockey agents,
competed in various sub-tasks (in simulation) and 
finished in full games (both in simulation and the real world).

When humans learn a new skill (such as air hockey),
it is common that a tutor/mentor will provide feedback and suggestions for their improvement.
The learner internalizes this feedback and augments their behavior in such a way to improve their future performance.
In contrast, current robot systems require experts to update a control policy directly to improve the model performance.
We are utilizing our framework as an intuitive interface to enable a human non-expert to modify the control policy of the robot in order to improve the systems' performance within the context of the Robot Air Hockey Challenge. 
The goal being to evaluate the low-level atomic action execution performance across diverse scenarios.
Experiments will be reported in future updates of this article.

\section{Conclusions}

In this section, 
we provide an overview of our main conclusions, 
acknowledge the current limitations of our framework and 
highlight potential future directions.

\subsection{Overview}

This work presents a ROS framework for intuitive robot programming, 
leveraging natural language prompts and contextual information from the robot environment and human feedback. 
Through a series of experiments conducted on a real-world robot setup, we have demonstrated the efficacy and versatility of our framework in enabling non-experts to program robots. 
The integration of open-source language models and common tools such as ROS with an AI agent represents a step towards realizing automated robotic solutions that can address real-world challenges in research and industry.
We have reported two case studies that indicate how our framework is demonstrably accelerating companies and research groups into useful deployment in different types of environments and scales.

\subsection{Limitations}

It is important to acknowledge the current limitations of our system. 
Presently, our experiments rely solely on action sequences, 
which are not adaptable without human feedback.
For example, if the robot must open a door and navigate through, then it may be worthwhile to attempt opening the door several times in case of failure of the first or subsequent attempts. 
This repetition could be achieved with a behavior tree, for example.
We have recently added an interface that is able to execute behavior trees generated by the language model in the form of an XML file - this feature is already usable in our code base.
Future work will investigate the benefits of using behavior trees in real-world laboratory setups.

Currently, we provide feedback, i.e. an observation, only in text form.
This neglects potentially informative data such as images and interaction forces with humans. 

Our framework provides a reward function in the problem formulation.
One limitation is that the reward function is heavily dependent on the failure flag.
In some cases of command, such as opening a door without specifying an angle, it can be difficult to determine whether the action is successful.
Also, the design of the reward function focused on achieving task-specific outcomes.
As tasks become more complex, involving a sequence of atomic actions, the dependency on a single reward signal at the end of the sequence may not provide enough guidance to the agent.


\subsection{Future work}

There are several avenues for future research and development we are interested in pursuing.
We plan to extend our framework to more versatile robotic platforms, including quadrupeds and quadrupeds with an arm. 
Additionally, we aim to integrate our system with vision-language models to enhance its perceptual capabilities.

Moreover, while our current system supports ROS 1 (Noetic), 
we intend to provide support for ROS 2 (Humble) to broaden its compatibility and applicability. 
Additionally, we plan to explore the use of other behavior representations, such as state machines and behavior trees, to further enhance the flexibility and adaptability of our framework. 

Whilst we have shown promise that pre-trained language models are capable of orchestrating action sequences for several tasks, there is an opportunity to fine-tune the language model.
However, there are significant challenges associated with fine-tuning language models for robotics, notably the issue of the gap between simulation and reality.  
We aim to investigate the potential of fine-tuning language models within our framework.

Finally, as mentioned in the previous sub-section, our reward function is limited. 
Future work will additionally investigate reward shaping for more diversified tasks, including nonbinary rewards.

The standardization and versatility of our framework in a variety of industries and research settings are creating opportunities for new collaborations, faster development, and propelling newly developed technologies forward. 
This trend will likely continue to manifest in the coming years as the ROS-LLM framework continues to mature.

In our approach, atomic actions are pre-trained or pre-defined policies. In future work, we may be able to fine-tune these individual policies to have faster and more effective atomic actions.

\bibliographystyle{plain}
\bibliography{bib.bib}

\end{document}